\documentclass{article}
\usepackage[margin=1.0in]{geometry}
\usepackage{hyperref}
\usepackage{amsmath}
\usepackage{algorithm, tabularx}
\usepackage[noend]{algpseudocode}
\usepackage{epsfig}
\usepackage{amssymb}
\usepackage{graphicx}
\usepackage{caption}
\usepackage{subcaption}
\usepackage{xcolor} 
\usepackage{array}
\usepackage{booktabs}
\usepackage{multirow}
\title{A Unified Framework for Multiclass and Multilabel Support Vector Machines}

\author{\Large Hoda Shajari \thanks{\small shajaris@ufl.edu} \qquad Anand Rangarajan}

\date{\large Dept. of Computer and Information Science and Engineering \\ University of Florida, Gainesville, FL}

\begin{document}

\maketitle
\begin{abstract}
 We propose a novel integrated formulation for multiclass and multilabel support vector machines (SVMs). A number of approaches have been proposed to extend the original binary SVM to an all-in-one multiclass SVM. However, its direct extension to a unified multilabel SVM has not been widely investigated. We propose a straightforward extension to the SVM to cope with multiclass \emph{and} multilabel classification problems within a unified framework. Our framework deviates from the conventional soft margin SVM framework with its direct oppositional structure. In our formulation, class-specific weight vectors (normal vectors) are learned by maximizing their margin with respect to an \emph{origin} and penalizing patterns when they get too close to this origin. As a result, each weight vector chooses an orientation and a magnitude with respect to this origin in such a way that it best represents the patterns belonging to its corresponding class.  Opposition between classes is introduced into the formulation via the minimization of pairwise inner products of weight vectors. We also extend our framework to cope with nonlinear separability via standard reproducing kernel Hilbert spaces (RKHS). Biases which are closely related to the origin need to be treated properly in both the original feature space and Hilbert space. We have the flexibility to incorporate constraints into the formulation (if they better reflect the underlying geometry) and improve the performance of the classifier. To this end, specifics and technicalities such as the origin in RKHS are addressed. Results demonstrates a competitive classifier for both multiclass and multilabel classification problems. 
\end{abstract}


\section{Introduction}\label{introduction}
Supervised classification is of great importance in many domains and research areas and has enjoyed widespread application. In a vast majority of cases, it is usually assumed that each pattern or instance (during training and testing) only belongs to one class and therefore is assigned a single label. While this kind of single label classification has been widely studied, there are many situations which do not fit into the single label classification framework with the classification problems being inherently multilabel. In the multilabel setting, an instance can simultaneously belong to multiple classes and therefore more than one label needs to be assigned to each instance. Consider the image tagging problem for instance. Here, each image contains multiple objects, each possessing a different label. The task at hand is the assignment of all relevant labels to each image comprising the set of unique objects in the scene. Alternatively, consider the problem of identifying each instrument at every time instant of a song. Since multiple instruments---guitar, bass and drums for example---can be simultaneously heard at any given moment, instrument recognition is inherently a multilabel classification problem. 

\subsection{Motivation}
Multilabel classification is a generalization of the well known multiclass problem in supervised learning and is consequently more challenging. Some of the challenges in the multilabel setting include, but are not limited to, exponential growth of cardinality of possible label sets, correlation among labels, structured label spaces and severely unbalanced datasets. Multilabel classification has already seen many real-world applications such as scene labeling \cite{boutell2004SceneSVM}, functional genomics analysis \cite{zhang2006multilabel,lippert2010gene} and text categorization \cite{joachims1998text,mccallum1999multi,kazawa2005maximal,rousu2006kerneltext}. 

The rising number of applications has led to the development of multilabel learning as a new supervised learning paradigm \cite{gibaja2015tutorial}. Many approaches have been proposed in the literature to tackle multilabel classification problems within the support vector machine (SVM) framework. Despite the plethora of approaches and algorithms, we assert that no straightforward (natural) extension of the SVM has hitherto been proposed for multilabel classification. The SVM was first introduced as a two-class, single label (binary) classifier based on maximum margin geometry (and regularized error minimization) \cite{boser1992training,cortes1995support,vapnik1998statistical}. Since then, the binary SVM has been studied extensively and applied successfully in many different domains. However, its extension to multiclass and multilabel cases is still an ongoing research problem since there is neither consensus nor a majoritarian approach.

In this work, we first attempt to justify the need to revisit the two-class (binary) SVM framework. Subsequently, we investigate the multilabel problem from a unified large margin perspective. We propose a novel formulation based on a geometric reinterpretation of the binary SVM which has a natural extension resulting in a unified multiclass and multilabel classification framework. Our proposed method is also capable of dealing with the aforementioned class imbalance problem in classification tasks. 

Multilabel classification approaches mainly fall into the two categories of problem transformation and algorithm adaptation. An intuitive way to handle multilabel classification tasks is to transform them into a series of binary classification problems and then use---as a method of solution---one of the existing binary classifiers (including binary SVM) to solve them. Algorithm adaptation approaches, on the other hand, solve the problem directly by extending binary classification methods in such a way that all classes are considered at once. Our method falls into the category of algorithm adaptation for multilabel (and multiclass) classification---within the SVM framework. In this paper, we therefore focus mainly on SVM-based approaches to multiclass and multilabel problems and mention other approaches (available in the literature) only as required.

\subsection{Outline of the basic idea}
Below, we describe the basic idea behind our multilabel SVM formulation. A more formal description is elaborated in Section~\ref{Unified SVM Classification Framework}. The main reason for presenting a pr\'{e}cis is to quickly get to the (simple) idea behind the formulation.

If the dataset is assumed to be linearly separable, then there are infinitely many hyperplanes which can separate the patterns into two classes. The main idea of the binary SVM is to choose a hyperplane which is optimal in the sense of maximizing the margin between the two sets of patterns while being somewhat robust to outliers. Given the finite set of patterns 
\[
(x_{1},y_{1}),(x_{2},y_{2}),\ldots,(x_{N},y_{N})\quad x_{i}\in\mathbb{R}^{K},\;y_{i}\in\{-1,1\},
\]  
subset $C_{1}\left(y=1\right)$ is linearly separable from subset $C_{2} \left(y=-1\right)$ by hyperplane $v^{T}x=c$ if there exist both a unit vector $v$ and constant c such that for all training instances, $y_{i}(v^{T}x_{i})>c$. Vapnik \cite{vapnik1998statistical} shows that the optimal solution $\left(v^{*}\right)$ to the following optimization problem is a unit normal vector of the hyperplane with maximum margin: 
\begin{equation}\label{Original Vapnik Formulation}
\begin{aligned}
    \max_{v}  &\quad\frac{1}{2}\left(\min_{x_{i}\in C_{1}}\;v^{T}x_{i}-\max_{x_{j}\in C_{2}}\;v^{T}x_{j}\right)\\
    \textup{s.t.}  &\quad\left\Vert v\right\Vert_{2}=1,
\end{aligned}
\end{equation}
where the assumption is that the patterns of these two classes are linearly separable, i.e., $y_{i}\left(v^{T}x_{i}\right)> c, i\in \left\{ 1,2,\ldots,N\right\}$ for some vector $v$ and constant $c$. The optimal unit vector $\left(v^{*}\right)$ along with constant
\begin{equation*}
c^{*}=\frac{1}{2}\left(\min_{x_{i}\in C_{1}}\;x_{i}^{T}v^{*}+\max_{x_{j}\in C_{2}}\;x_{j}^{T}v^{*}\right),
\end{equation*} 
determine the maximum margin hyperplane. It is also proved that the maximum margin (optimal) hyperplane is unique \cite{vapnik1998statistical}. Furthermore, the optimization problem in (\ref{Original Vapnik Formulation}) can be equivalently formulated as finding a pair---vector $w$ and bias (threshold) $b$--such that $w$ has the smallest $\ell_{2}$ norm and constraints $y_{i}\left(w^{T}x_{i}\right)+b\geq1$ are satisfied for all training patterns \cite{vapnik1998statistical}:
\begin{equation}\label{Equivalent Vapnik Formulation}
\begin{aligned}\min_{w,b} & \quad\frac{1}{2}\left\Vert w\right\Vert _{2}^{2}\\
\textup{s.t.} & \quad y_{i}\left(w^{T}x_{i}+b\right)\geq 1,\quad i\in\left\{ 1,2,\ldots,N\right\}.
\end{aligned}
\end{equation}
Patterns for which the inequality constraints in (\ref{Equivalent Vapnik Formulation}) become active (equality constraints) are of particular importance. These points locate the optimal hyperplane and are known as support vectors and hence the name support vector machine. Cortes and Vapnik \cite{cortes1995support} extended this problem to the case where the two classes are not linearly separable by relaxing the hard margin constraints in (\ref{Equivalent Vapnik Formulation}) and letting some patterns violate the margin:
\begin{equation}\label{slack included formulation}
    \begin{aligned}\min_{w,\left\{ \xi_{i}\right\} ,b} & \quad\frac{1}{2}\left\Vert w\right\Vert _{2}^{2}+C\sum_{i}\xi_{i}\\
\textup{s.t.} & \quad y_{i}\left(w^{T}x_{i}+b\right)\geq1-\xi_{i},\quad i\in\left\{ 1,2,\ldots,N\right\} \\
 & \quad\xi_{i}\geq0, \quad i\in\left\{ 1,2,\ldots,N\right\}.
\end{aligned}
\end{equation}
The optimization problem in (\ref{slack included formulation}) is known as the soft margin SVM. Solving for $\{\xi_{i}\}$ in terms of the other variables, we get the following \textit{hinge loss} in objective function which is an upper bound convex approximation to the zero-one misclassification loss function:
\begin{equation*}
\min_{w,b}\quad\frac{1}{2}\left \| w\right \|_{2}^{2}+C\sum_{i}\left[1-y_{i}\left(w^{T}x_{i}+b\right)\right]_{+}.
\end{equation*}
The essential idea behind our multilabel (and multiclass) formulation was inspired by the optimization problem in (\ref{Original Vapnik Formulation}) which can be reformulated as
\begin{equation*}
    \begin{aligned}\max_{v} & \quad\frac{1}{2}\left(\min_{x_{i}\in C_{1}}\;v^{T}(x_{i}-x_{0})-\max_{x_{j}\in C_{2}}\;v^{T}(x_{j}-x_{0})\right)\\
\textup{s.t.} & \quad\left\Vert v\right\Vert _{2}=1
\end{aligned}
\end{equation*}
and furthermore, equivalently written as
\begin{equation}\label{twinmargin}
\begin{aligned}
\max_{v} & \quad\frac{1}{2}\left(\min_{x_{i}\in C_{1}}\;v^{T}(x_{i}-x_{0})+\min_{x_{j}\in C_{2}}\;(-v)^{T}(x_{j}-x_{0})\right)\\
\textup{s.t.} & \quad\left\Vert v\right\Vert _{2}=1.
\end{aligned}
\end{equation}
We have introduced an origin $x_{0}$ and wish to point out that reconfiguring the training set with respect to this origin does not change the optimal solution to the problem in (\ref{Original Vapnik Formulation}). In fact, Vapnik's original formulation in (\ref{Original Vapnik Formulation}) is relative. Therefore, adding $\left(v^{T}x_{0}\right)$ to both terms in the optimization problem in (\ref{Original Vapnik Formulation}) does not change the optimal solution. In Vapnik's equivalent formulation (\ref{Equivalent Vapnik Formulation}), the origin is implicitly introduced via a bias term $b$ by transitioning from the relative formulation in (\ref{Original Vapnik Formulation}). Furthermore, we are effectively considering a separate hyperplane for each class which results in the two hyperplanes becoming parallel in this formulation with the addition of a sum-to-zero constraint on the normal vectors $(v+(-v)=O)$. As we will see in Section~\ref{Unified SVM Classification Framework}, there is no necessity to have a sum-to-zero constraint on the normal vectors (of hyperplanes). When this is relaxed, it yields \emph{nonparallel} separating hyperplanes in two-class problems. While it may seem odd to have twin, nonparallel hyperplanes in a two-class problem, as we later show, it becomes a useful construct in the multilabel setting.

Having presented the basic idea---nonparallel hyperplanes---we turn to a quick description of related work in order to forestall impressionistic notions of similarity to previous work. The twin SVM (TWSVM) also departs from parallel hyperplanes for the two-class problem \cite{khemchandani2007twin}. However, it is neither a maximum margin formulation nor an all-in-one machine since it solves two separate optimization problems for binary classification. Mangasarian and Wild \cite{mangasarian2005multisurface} proposed a generalized eigenvalue proximal SVM (GEPSVM) which deviates from a previous proximal SVM (PSVM) \cite{Mangasarian2005multicategory} with parallel hyperplanes. This approach also solves two separate optimization problems for obtaining the separating hyperplane of each class in binary classification. The normal vector of each of the two nonparallel proximal hyperplanes is the eigenvector corresponding to the smallest eigenvalue of a generalized eigenvalue problem. 

Our approach is also different from the one-class SVM which is an unsupervised algorithm for outlier (novelty) detection \cite{scholkopf2000support,tax2004support}. In the one-class SVM, training data is available for only one class and the goal is to decide whether a new pattern belongs to this class. The one-class SVM introduced in \cite{scholkopf2000support} is a natural extension of SVM to the case of unsupervised learning and all the patterns are separated from the origin of the feature space by a hyperplane with maximum margin. Depending on which side of the hyperplane a point falls, it is classified either as a member (of that class) or as an outlier. Whether or not this problem could be formulated for a two-class problem with outliers without the oppositional framework has not been investigated. This approach does not consider the multilabel problem (or even multiclass for that matter). Our model simultaneously addresses multiclass and multilabel problems with the origin implicitly determined by the support vectors of all classes in feature space. In particular, our problem formulation is based on a unique and novel geometry for simultaneously maximizing the margins of all classes in the multiclass and multilabel settings. In order to better demonstrate the advantages of our approach, we mainly focus on large margin formulations in the literature. 

The outline of paper is as follows. In Section~\ref{related works} we review previous work focused on multiclass and multilabel problems mostly within the SVM framework. In Section~\ref{Unified SVM Classification Framework}, the new formulation is presented and a quadratic majorizer for the hinge loss is used to solve the optimization problem in the primal. In Section~\ref{Generalization to RKHS}, the nonlinear transformation of the input space to infinite dimensional spaces via reproducing kernel Hilbert spaces (RKHS) is discussed and the majorization algorithm extended to solve the primal kernel formulation. In Section~\ref{experiments}, we present illustrative experiments and the results of our formulation on multiclass and multilabel datasets. We also discuss a scenario which widely used approaches are not able to handle but is easily handled in our formulation. Section~\ref{Conclusion} presents the conclusions and speculates on future possibilities and extensions.

\section{Related Work}\label{related works}
In our discussion of previous work, we mainly confine ourselves to SVM-based approaches, specifically all-in-one (single) machines for multiclass and multilabel problems. All-in-one approaches consider information of all classes simultaneously and set up a single optimization problem for learning all classifiers (discriminators) at once. The margin concept is key in binary SVM and its extensions. SVM approaches differ mainly in how they define the margin and how margin violations are formulated in the loss function. There are two notions of margin, namely, relative margin and absolute margin which basically correspond respectively to cases where the margins of the classifiers are optimized relative to each other or otherwise \cite{dogan2016unified}.

Vapnik \cite{vapnik1998statistical} proposed the first all-in-one framework for the multiclass SVM based on relative margins. \cite{weston1999support} and \cite{bredensteiner1999multicategory} also used this framework albeit with slightly different formulations. It can be shown that all three formulations are equivalent. In this framework, the relative margin of each pattern is maximized against all other classes to which it does not belong with a slack variable utilized for each of these pairs. Crammer and Singer \cite{crammer2001algorithmic} proposed a similar formulation for multiclass SVM with a single slack variable per pattern, essentially a penalty for the worst class. Their formulation did not take the biases into account. Hsu and Lin \cite{hsu2002comparison} subsequently added biases into the Crammer and Singer formulation. 

Lee et al. \cite{lee2004multicategory} proposed another formulation for multiclass SVM with absolute margins which (while extending the Bayes decision rule for multiclass in the same fashion as for the binary case) ensures that the solution directly targets the Bayes decision rule also known as Fisher consistency. However, setting up the framework for compatibility with the Bayes decision rule eliminates the notion of support vectors for each class since the hinge loss term corresponding to the correct class gets eliminated. In other words, this approach penalizes a pattern if it gets within $\frac{1}{K-1}$ margin distance to other classifiers (where $K$ is the number of classes) without encouraging a pattern to get close to its own classifier. The work proposed by \cite{lee2004multicategory}, and \cite{weston1999support} are theoretically sound but an efficient training algorithm has not yet been proposed \cite{dogan2016unified}. Van Den Burg and Groenen \cite{van2016gensvm} introduced a single machine---the generalized multiclass SVM (GenSVM)---which uses a simplex encoding to cast the multiclass problem in $K-1$ dimensions. GenSVM generalizes aforementioned methods of \cite{lee2004multicategory} and \cite{crammer2001algorithmic}. Liu and Yuan \cite{liu2011reinforced} proposed a new type of multiclass hinge loss function called reinforced hinge loss and proved that under some conditions, it is Fisher consistent. 
 Szedmak et al. \cite{szedmak2006learning} introduced the multiclass maximum margin regression (MMR) framework with computational complexity no more than a binary SVM. They extend the idea of support vector regression \cite{vapnik1998statistical} to vector label learning in an arbitrary Hilbert space. Dogan et al. \cite{dogan2016unified} showed that most of the proposed all-in-one multiclass SVMs are mainly different in terms of choice of margin and margin-based loss functions and presented a unified formulation which accommodates existing single machine approaches for multiclass classification. Based on this unified view, they proposed two new multiclass formulations. 

Zhang and Jordan \cite{zhangJordan2012bayesian} proposed a Bayesian multiclass SVM by extending the method proposed in \cite{sollich2002bayesian}. \cite{Mangasarian2005multicategory} proposed multicategory proximal support vector machines (MPSVM) where instead of working in the feature space $\mathbb{R}^{n}$ and maximizing the margin, they concatenate the bias to the weight vector and try to maximize this extended margin in $\mathbb{R}^{n+1}$ \cite{Fung2001ProximalSV}. MPSVM is a problem transformation scheme which solves $K$ independent nonsingular systems of linear equations to obtain the weight vectors $w_{k},k\in\left\{1,\ldots, K\right\}$. It is not clear why merging the bias with the weight vector and maximizing this extended margin should improve classification performance.

The one versus the rest (OvR) a.k.a. one versus all (OvA) approach is based on the idea of separately training binary classifiers for each class. The patterns belonging to one class form a group and all other patterns form the second group. Therefore, in the OvR approach for $K$-class problems, $K$ binary classifiers are trained \cite{vapnik1998statistical}. For a test pattern, each classifier outputs a real value which reflects the level of certainty of the pattern belonging to its corresponding class. Naturally, the test pattern is assigned to the class with maximum discriminant function value or confidence score (winner-take-all decision criterion). OvR does not take label correlations into account. OvR is considered to be an intuitive and simple extension to binary SVM and yet powerful in producing results as good as other classifiers \cite{rifkin2004defenseOvA} and can be extended to the multilabel problem by using a probabilistic scheme for obtaining confidence levels and considering a threshold for scores for which each value or probability gives rise to a label. However, as we will show in Section~\ref{experiments}, there are cases in which this approach fails. Since a smaller number of patterns are used to train the classifier, OvR is susceptible to variance increase since partitioning data this way makes the training set unbalanced \cite{lee2004multicategory}. In the OvR approach, it is not clear how support vectors should be determined and their interpretation is ambiguous especially in the multilabel setting since support vectors are relative.

In the one versus one (OvO) approach a.k.a. all versus all, a binary classifier is trained for each pair of classes \cite{hastie1998OvO,kressel1999pairwise}. Therefore for a $K$-class problem, $K(K-1)/2$ classifiers are trained. OvO suffers from the transitivity problem which may result in label contradiction in some areas of the input space. A test pattern is assigned to the class with maximum number of votes. When the number of classes is large, these binary-classifier-based methods may suffer from either computational cost or highly imbalanced sample sizes in training. DAGSVM was proposed to combine the result of classifiers obtained via the OvO strategy \cite{platt2000DAGSVM}. A new test pattern is classified using $K-1$ classifiers which are chosen based on which path is traversed in the graph.

Dietterich  and  Bakiri \cite{dietterich1994error-correcting} proposed a general approach for multiclass classification based on error-correcting output codes (ECOC) for binary classifiers. A binary string of some fixed length $s$ known as the codeword is assigned to each class and then $s$ binary functions are learned. At the decoding step, a code is obtained for a test pattern in the test set by evaluating all trained binary classifiers for that pattern. The test pattern is assigned to the class with minimum Hamming distance (which counts the number of bits that differ) between the binary output code of the test pattern and base code of that class. ECOC is applied to margin-based methods including SVM where each codeword belongs to $\{-1,0,1\}^{s}$ with $0$ indicating that the way in which the corresponding classifier assigns this label is irrelevant \cite{allwein2000errorcoding}. Label prediction is based on a loss-based decoding scheme. Loss-based decoding takes the output magnitude of binary classifiers which are considered as certainty level into account and choose a label which is most consistent with the classifiers' outputs based on the loss function. Most of the aforementioned approaches do not accommodate and cannot be extended to multilabel classification in a straightforward manner. 

Approaches to multilabel learning mainly fall into the categories of algorithm adaptation and problem transformation. In general, there is no established approach for solving a multilabel classification problem \cite{scikit-multilearn}. OvR a.k.a binary relevance (BR) in multilabel learning is a baseline approach and usually serves as a benchmark for other multilabel approaches \cite{luaces2012binary}.   Boutell et al. \cite{boutell2004SceneSVM} used the OvR approach with SVM as the binary classifier on the scene dataset. They use cross training which basically includes each pattern as a positive example (label 1) for each class when training the corresponding classifier. The multilabel patterns are not used as negative examples while training the classifiers. In our framework, on the other hand, no patterns are used as negative examples for training any classifier since each classifier only uses information related to its own patterns to determine the hyperplane. They have also offered an empirical (heuristic) decision criterion for testing and an evaluation metric for performance evaluation and this is mentioned in Section~\ref{experiments}. Two approaches are proposed in \cite{godbole2004discriminative} to improve the margin in multilabel classification within the OvR framework. The first approach eliminates patterns from the opposite class (the rest). The second approach removes all the training data in confusing classes using a confusion matrix obtained from applying a fast and relatively accurate classifier. Chen et al. \cite{chen2016mltsvm} extended TWSVM to the multilabel setting and proposed MLTSVM in which $K$ classifiers are trained separately in OvR fashion.

Elisseeff and Weston \cite{elisseeff2002rankSVM} introduced a large margin ranking system known as Rank-SVM for multilabel classification. Rank-SVM is an algorithm adaptation method which can be considered as an extension of the Crammer and Singer multiclass SVM framework. The idea of Rank-SVM is to find $K$ classifiers (hyperplanes), one for each label, such that the relevant labels for each training pattern are ranked higher than irrelevant labels. To achieve this, a proposed ranking loss is minimized and relative margins are maximized. Since it is a ranking-based approach, the design of a set size predictor is needed. Formulating multilabel classification as a ranking problem imposes a quadratic number of constraints on the problem and makes optimization more challenging \cite{hariharan2010large}. Rank-SVM, as its name indicates, ranks the labels for each pattern and then chooses a number of highly ranked labels for each pattern based on a probabilistic or heuristic approach for determining the cardinality of the set of labels. Therefore, Rank-SVM is not able to directly produce output labels.

Armano et al. \cite{armano2012error} extended ECOC to multilabel text categorization. They generate a codeword for each class with only $1$ and $0$ bits. Then the posterior probability of each class is obtained and $t$ top ranking categories are selected where $t$ is user defined or tuned by the validation set. 

Incorporating label space structure and capturing label correlations within the framework of large margin methods has been studied in multilabel learning \cite{tsochantaridis2004structuredSVM,liu2018svm,hariharan2010large,sun2011canonical,lampert2011maximum,guo2011adaptive,taskar2004max}. There are cases in which elements of the output space are structured objects such as trees, sequences, strings or sets. The challenge here is how to effectively use label correlation information to improve accuracy in multilabel prediction. The structured SVM (SSVM) was proposed to learn a mapping from the input space to structured output spaces like a parse tree in Natural Language Parsing \cite{tsochantaridis2004structuredSVM,joachims2009predicting}. 

Hariharan et al. \cite{hariharan2010large} developed a maximum margin multilabel formulation (M3L) which assumes labels are densely correlated but did not include pairwise label correlation terms in the objective function. Their work is the middle ground between approaches with independent labels assumption (linear complexity) and explicitly modeling label correlations (exponential complexity). They make the assumption that prior knowledge about label correlations is available and that labels have at most linear dense correlations. This prior knowledge is incorporated into the formulation in the form of a dense correlation matrix and $K$ densely correlated sub-problems are solved. They show that OvR is obtained as a special case of their model. 

Canonical correlation analysis (CCA) is used for dimensionality reduction while exploiting label correlations in the multilabel setting \cite{sun2011canonical, Jiepeng2016book}. In supervised dimensionality reduction via CCA, the two sets of variables are derived from the data and class labels followed by projection into a lower-dimensional space in which the patterns are maximally correlated. They incorporated a dimensionality reduction scheme into the SVM formulation and proposed a joint dimensionality reduction and multilabel classification framework.
Their framework also uses opposition from other classes to learn classifiers. Liu et al. \cite{liu2018svm}, modified SVM to take missing labels into consideration by extending the label set to include zero which indicates a missing class label. They also take label correlations into consideration by proposing label and class smoothness and integrating these into the objective function. 

These approaches still use patterns from other classes in an oppositional fashion to obtain each classifier. Our approach is a natural extension to an all-in-one  multiclass SVM \cite{weston1999support,crammer2001algorithmic} and does not need to incorporate every possible label set prediction into model. Labels get assigned only after training---similar to multiclass classification in SVM. Furthermore, interpretation of labels is very important in multilabel classification problems since in most of the work, not having a label for a class is automatically interpreted as negative membership. Our framework takes this into account and uses available information on memberships for classification and deviates from all oppositional classification frameworks. However, it is easily extensible to negative labels by assigning a label of $-1$ to patterns for which information about negative membership is available. Our approach resides between approaches with the independent labels assumption and the ones which consider direct pairwise label correlations. From this perspective, we do not have the exponential label space complexity issue since separation in our framework is against the origin.  As a matter of fact, our formulation does not explicitly explore label correlations, however, as we show, this information is implicitly taken into account. Finally, a dimensionality reduction scheme can also be easily incorporated into our formulation.

\section{A Unified SVM Classification Framework}\label{Unified SVM Classification Framework}
The underlying motivation for the present work originates from the desire to naturally extend the SVM formulation to a multilabel classification framework. The lack of such a unified multiclass and multilabel framework in the literature led us to revisit the original SVM formulation and suggest a new, geometrically driven approach. The basic idea is as follows. We (implicitly) learn a new origin for the feature vectors as well as the magnitude and direction of the weight vectors $(\left\{ w_{k}\right\})$ such that the inner product of the reconfigured patterns with respect to this new origin $(x_{i}-x_{0})$ with each weight vector $w_{k}$ represents the confidence (certainty) of the memberships of the patterns belonging to each class. Consequently, each class $C_{k}$ chooses a magnitude and orientation of its weight vector $w_{k}$ in such a way that the patterns belonging to $C_{k}$ are best represented (separated) without the need of explicit opposition to other classes. Inspired by this characteristic, our approach is called \emph{one versus none (OvN)}. In this section, we flesh out this idea, formulate a new objective function and present an algorithm for finding the global optimum.

\subsection{Problem Formulation}
In Section~\ref{introduction}, we motivated the unified framework by jettisoning the parallel hyperplane constraint which is standard in almost all binary (two-class) SVMs. Furthermore, we introduced an origin variable $x_{0}$ and set up the margin constraints using this new variable. The picture that emerges from this description is of individual class-specific weight vectors $(\left\{ w_{k}\right\})$ which seek to satisfy margin constraints for the patterns belonging to that class. This is repeated for all classes. However, this alone does not ensure competition between classes which is enshrined using parallel hyperplanes in the standard SVM. In our formulation, we introduce competition by minimizing the inner product between all pairs of weight vectors. Note that the minimum inner product between two unit weight vectors $w_{k}$ and $w_{l}$ is attained when the two vectors are anti-parallel (as in the standard SVM). Later, in Section~\ref{binary_svm_equivalance}, we show that suitable constraints on a pair of weight vectors $w_{1}$ and $w_{2}$ in a two-class problem, make our formulation completely equivalent to the standard binary SVM.

In common with virtually all previous soft margin approaches, the objective function includes regularization and outlier rejection terms \cite{cortes1995support}. As mentioned above, in addition to the margin term for each weight vector $w_{k}$, we have the sum of all pairwise inner products between weight vectors $w_{k}$ and $w_{l}$ which endeavor to make the weight vectors anti-parallel. Thus far, we have not discussed the bias term in this set up. When we introduced the origin $x_{0}$ as a variable in (\ref{twinmargin}), we ended up with two terms that canceled each other out: $v^{T}x_{0}$ and $-v^{T}x_{0}$ which assumes anti-parallel hyperplanes. In a two-class multilabel SVM, these become $b_{1}\equiv -w_{1}^{T}x_{0}$ and $b_{2}\equiv -w_{2}^{T}x_{0}$. We then have the option of enforcing a hard constraint $b_{1}+b_{2}=0$ or a soft constraint term $\left(b_{1}+b_{2}\right)^{2}$ to be mimimized as part of objective function. For problem formulation with more than two classes, the former option is selected in this work since it was practically demonstrating better performance in both linear and kernel cases and is fully demonstrated in appendices. The other case can be similarly worked out. From a conceptual perspective, a hard constraint on the biases is an extension of the anti-parallel hyperplanes constraint, whereas the soft constraint on the biases is a relaxation of the same. A candidate multilabel (and multiclass) optimization problem is 
\begin{equation}\label{main formulation}
\begin{aligned}\min_{\left\{ w_{k}\right\} ,\left\{ \xi_{ki}\right\} ,x_{0}} & \quad\frac{1}{2}\sum_{k=1}^{K}\parallel w_{k}\parallel_{2}^{2}+\alpha\sum_{k=1}^{K}\sum_{l=k+1}^{K}w_{k}^{T}w_{l}+\beta\sum_{k=1}^{K}\sum_{i\in C_{k}}\xi_{ki}\\
\textup{s.t.}\quad\; & \quad\sum_{k=1}^{K}\left(-w_{k}^{T}x_{0}\right)=0\\
\textup{} & \quad w_{k}^{T}(\,x_{_{ki}}-x_{0})\geq1-\xi_{ki},\quad\forall k,\quad\forall i\in C_{k}\\
 & \quad\xi_{ki}\geq0,
\end{aligned}
\end{equation}
where the first constraint is the hard constraint on biases and the second set of constraints are margin violation constraints obtained by introducing outlier (slack) variables \cite{cortes1995support}. The first term in the objective function relates to margin maximization, the second term implicitly models pairwise label correlations and the last term penalizes deviation of patterns from their corresponding class margin. The regularization parameters, $\alpha$,$\;\beta>0$, have standard objective function semantics. We reformulate the optimization problem in (\ref{main formulation}) by eliminating all the outlier variables via hinge losses and introducing the bias variables $b_{k}\equiv-w_{k}^{T}x_{0}$:
\begin{equation}\label{main-replaced-by-b}
\begin{aligned}\min_{W,b} & \quad\frac{1}{2}\sum_{k=1}^{K}w_{k}^{T}w_{k}+\alpha\sum_{k=1}^{K}\sum_{l=k+1}^{K}w_{k}^{T}w_{l}+\beta\sum_{k=1}^{K}\sum_{i\in C_{k}}\left[1-\left(w_{k}^{T}x_{ki}+b_{k}\right)\right]_{+}\\
\textup{s.t.}& \quad\sum_{k=1}^{K}b_{k}=0,
\end{aligned}
\end{equation}
where we define $W = \left\{w_{1},w_{2},...,w_{K}\right\}$ as the set comprising all the normal vectors to the hyperplanes and  $b=\left\{b_{1}, b_{2},...,b_{K}\right\}$ as the set of biases. The hard constraint on biases is incorporated into the objective function of (\ref{main-replaced-by-b}) by the method of Lagrange multipliers and results in the following Lagrangian (where $\lambda$ is a Lagrange multiplier):
\begin{equation}\label{Lagrangian function for linear case}
\mathcal{L}\left(W,b,\lambda\right)=\frac{1}{2}\sum_{k=1}^{K}w_{k}^{T}w_{k}+\alpha\sum_{k=1}^{K}\sum_{l=k+1}^{K}w_{k}^{T}w_{l}+\beta\sum_{k=1}^{K}\sum_{i\in C_{k}}\left[1-\left(w_{k}^{T}x_{ki}+b_{k}\right)\right]_{+}+\lambda\sum_{k=1}^{K}b_{k}.    
\end{equation}
This is by no means the only available integrated formulation. In Section~\ref{linear soft and hard constraints}, we examine all four possibilities: the cross product space of soft and hard constraints on the hyperplanes and soft and hard constraints on the biases. We also have considered the margins as in the conventional SVM. However, it is possible to incorporate margin-related variables in the formulation by setting up the problem similar to the $\nu$-SVC \cite{scholkopf2000new}.
\subsection{Testing Scenario}
After solving the optimization problem and obtaining the classifiers, a test instance is projected to each weight vector. We employ a \emph{simple} decision criterion wherein each test instance gets assigned all class labels with projection greater than $1$, i.e., if $S_{t}$ denotes the set of labels for test instance $x_{t}$, then
\[{S_{t}=\{k:w_{k}^{T}x_{t}+b_k\geq1\}}.
\]
From a geometric point of view, we have an ambiguous region which is inhabited by  patterns whose projections are all less than 1. Therefore the decisions on label assignment could include this information. In the present work however, we eschew careful examination of pattern and label test set geometry and relegate this to future work. Instead, when this situation occurs, we revert to a multiclass strategy and pick the label corresponding to the maximum projection as explained below.

In the multiclass (single label) setting, a winner-take-all strategy (as mentioned above) is applied and the test pattern is assigned to the class with maximum projection value 
\[
m=\underset{k}{\textup{arg\,max}}\:\{w_k^T x_t + b_k\}.
\]
\subsection{A Quadratic Majorizer for Optimization}\label{A Quadratic Majorizer for Optimization}
Most of the existing literature on SVM optimization is focused on formulating the primal problem from which the dual formulation is extracted followed by a solution in dual space. However, primal problems can also be efficiently solved. In fact, solving the primal may have advantages for large scale optimization since---at least in the linear SVM---the objective function is linear in the training set patterns (and not quadratic). Furthermore, often in large scale optimization, an approximate solution is sought which results in a large number of support vectors. In this case, the dual may result in a solution which is not meaningful \cite{chapelle2007primalSVM}. 

One of the most simple and effective algorithms for solving the primal is majorization-minimization (MM). The main advantage of iterative majorization for the SVM is that it results in simple linear system updates (achieved without the need for line-search parameter tuning). A secondary advantage of MM is that the objective function value is guaranteed to be non-increasing and is generally decreasing in each iteration. If the function to be optimized is strictly convex which is the case with the SVM loss function, MM converges to the global minimum. In contrast, other methods which solve the dual must often completely converge in order to obtain a meaningful and reasonable solution. Variants of MM have also been proposed which deal with large-scale optimization problems. 

Consider the standard, linear, binary SVM in the primal. It comprises a margin term and the hinge loss on the patterns. An alternative to majorization is to smooth the hinge loss (resulting in a differentiable approximation like the Huber loss). However, in this case, we are minimizing a different function which is an approximation to the original function. Furthermore, the smoothed hinge loss does not result in simple update equations (achieved without having to estimate a step-size parameter). Even if we approximate the hinge loss with the generalized logistic loss $g_{\beta}(x)=\frac{1}{\beta}\text{log}(1+\text{exp}\{\beta(1-x)\})$ (which approaches $[1-x]_{+}$ as $\beta\rightarrow\infty$), we would still need to use nonlinear optimization on this logistic regression variant in order to obtain good solutions. Finally, the non-differentiability of the hinge loss forces us to adopt subgradient optimization which is also complex (from a line search perspective). For these reasons, we introduce a simple MM approach to minimize the SVM objective in our unified framework. Clearly, alternatives are available (including minimization in the dual) but we have relegated these to future work.

Majorization leads to a particular kind of iterative optimization algorithm. The basic idea behind majorization is to construct an auxiliary objective function which lies above the original objective function and is easier to minimize. The auxiliary objective function usually has the property of touching the original objective at the current time (iteration) instant. By descending on the auxiliary objective, we guarantee that we also descend on the original objective by virtue of the fact that the auxiliary objective is strictly greater than or equal to the original \cite{lange2013optimization}. For the task at hand, note that MM can specifically deal with nondifferentiable convex objective functions (like the hinge loss) by constructing a sequence of smooth convex functions which majorize the original nonsmooth objective function. A brief formal description of MM follows. A function $g(x,x_{k})$ majorizes $f\left(x\right)$ at $x_{k}$ if
\begin{equation}\label{majorization conditions}
    \begin{aligned}f\left(x_{k}\right) & =g\left(x_{k,},x_{k}\right),~\mathrm{and}\\
f(x) & \leq g\left(x,x_{k}\right),\quad x\neq x_{k},
\end{aligned}
\end{equation}
which means that $g(x,x_{k})$ lies above $f\left(x\right)$ everywhere and touches $f(x)$ at least at $x_{k}$ where $x_{k}$ is a point in the domain of $f(x)$ at the current iteration. Now we minimize the surrogate function $g\left(x,x_{k}\right)$ instead of $f\left(x\right)$ in the minimization version of the MM algorithm. If $x_{k+1}$ is the minimum of $g\left(x,x_{m}\right)$, then the descent property follows from (\ref{majorization conditions}). From the descent property, we see that it is possible to merely descend on $g$ instead of finding its global minimum:
\[
f\left(x_{k+1}\right)\leq g\left(x_{k+1},x_{k}\right)\leq g\left(x_{k},x_{k}\right)=f\left(x_{k}\right).
\]

There exist previous approaches which solve the SVM in the primal using MM
\cite{groenen2007majorize,SVM-Maj-2008,nguyen2017iterativelyS,van2016gensvm}. For example, \cite{groenen2007majorize} have a majorization approach which considers different loss functions (hinge, quadratic, Huber). We follow \cite{lange2009sharp} in our approach. \cite{groenen2007majorize} and \cite{lange2009sharp} show that the hinge loss $[1-u]_{+}:=\textup{\ensuremath{\min}}(0,1-u)$ is majorized by the following quadratic (in $u$):
\begin{equation*}\label{guzhinge} 
    g(u,z)=\frac{1}{4z}(1-u+z)^{2}
\end{equation*}
where $z>0$. It is shown that $g$ is the best quadratic majorizer for $\left[1-u\right]_{+}$ \cite{lange2009sharp}. Theoretically, it is possible to get arbitrarily as close as possible to the hinge loss through the above quadratic majorizer. In practice however, we choose a very small positive threshold value $\epsilon$ for $z$, i.e., $z\geq\epsilon$ for the sake of numerical stability. Here $z$ is an auxiliary variable. Assuming (for the sake of convenience) that $z$ is unconstrained (instead of requiring a positivity constraint), the derivative of $g(u,z)$ w.r.t. $z$ is
\[
\frac{\partial g}{\partial z}=\frac{1}{4}-\frac{(1-u)^{2}}{4z^{2}}.
\]
Setting this to zero yields $z^{2}=(1-u)^{2}$ from which we obtain $z=\left|1-u\right|$ which is non-negative and is equal to zero when $u=1$. It can easily be shown that setting $z=\epsilon$ when $|1-u|<\epsilon$ is the global minimum for $z$ (when $z$ is constrained to be greater than or equal to $\epsilon$). We can therefore use the above majorization trick to obtain a quadratic objective w.r.t. $u$. Note that $g(u,z)$ will be replaced by a suitable function of $w_{k}$ and $b_{k}$ in the formulation. The majorized multiclass/multilabel SVM optimization problem can be written as
\begin{equation}\label{main-majorizer}
\begin{aligned}\min_{W,b,Z} & \quad\frac{1}{2}\sum_{k=1}^{K}w_{k}^{T}w_{k}+\alpha\sum_{k=1}^{K}\sum_{l=k+1}^{K}w_{k}^{T}w_{l}+\beta\sum_{k=1}^{K}\sum_{i\in C_{k}}\overset{\textup{hinge\;loss\;majorizer}}{\overbrace{\frac{1}{4z_{ki}}\left[1+z_{ki}-(w_{k}^{T}x_{ki}+b_{k})\right]^{2}}}\\
\textup{s.t.}& \quad\sum_{k=1}^{K}b_{k}=0,
\end{aligned}
\end{equation}
where $Z$ denotes the set of all strictly positive auxiliary variables $\left\{z_{ki}\right\}$. In order to simplify the notation in the optimization problem in (\ref{main-majorizer}), we concatenate $w_{k}$ and $b_{k}$ into one vector and define the following matrices:
\begin{equation}\label{matrices soft w hard b}
    \widetilde{w}_{k}\equiv\left[\begin{array}{c}
w_{k}\\
b_{k}
\end{array}\right],\;\widetilde{x}_{ki}\equiv\left[\begin{array}{c}
x_{ki}\\
1
\end{array}\right],\;D\equiv\left[\begin{array}{cc}
I_{M\times M} & \overrightarrow{0}\\
\overrightarrow{0}^{T} & 0
\end{array}\right],\quad e\equiv\left[\begin{array}{c}
0\\
\vdots\\
0\\
1
\end{array}\right]_{(M+1)}
\end{equation}
where $M$ is the number of features. With these changes and defining the set of augmented weight vectors $\widetilde{W}=\left\{\widetilde{w}_{1},\ldots,\widetilde{w}_{K}\right\}$, we may rewrite the optimization problem as
\begin{equation}
\begin{aligned}\min_{\widetilde{W}, Z} & \quad\sum_{k=1}^{K}\widetilde{w}_{k}^{T}D\widetilde{w}_{k}+\alpha\sum_{k=1}^{K}\sum_{l=k+1}^{K}\widetilde{w}_{k}^{T}D\widetilde{w}_{l} +\beta\sum_{k=1}^{K}\sum_{i\in C_{k}}\frac{1}{4z_{ki}}\left[1+z_{ki}-(\widetilde{w}_{k}^{T}\widetilde{x}_{ki})\right]^{2}\\
 \textup{s.t.}&\quad \sum_{k=1}^{K}e^{T}\widetilde{w}_{k}=0,
\end{aligned}\label{augmented majorized SVM formulation}
\end{equation}
and again by including the hard constraint on biases into the objective function via the method of Lagrange multipliers, the Lagrangian can be written as
\begin{equation}\label{Lagrangian function linear case}
\begin{aligned}\mathcal{L}(\widetilde{W},Z,\lambda)= & \sum_{k=1}^{K}\widetilde{w}_{k}^{T}D\widetilde{w}_{k}+\alpha\sum_{k=1}^{K}\sum_{l=k+1}^{K}\widetilde{w}_{k}^{T}D\widetilde{w}_{l}\\
 & +\beta\sum_{k=1}^{K}\sum_{i\in C_{k}}\frac{1}{4z_{ki}}\left[1+z_{ki}-(\widetilde{w}_{k}^{T}\widetilde{x}_{ki})\right]^{2}+\lambda\sum_{k=1}^{K}e^{T}\widetilde{w}_{k}.
\end{aligned}
\end{equation}
Rather than solving for $\widetilde{W}$, we concatenate all the augmented weight vectors $\widetilde{w}_{k}$ into one vector $\widetilde{w}=\left[\widetilde{w}_{1}^{T},\widetilde{w}_{2}^{T},\ldots,\widetilde{w}_{K}^{T}\right]^{T}$ and alternate between least-squares solutions for $\widetilde{w}$ and closed-form solutions for $Z$. The solution for $Z$ follows the approach outlined above:
\begin{equation}\label{z update}
     z_{ki}=\begin{cases}
z_{ki} & \left|1-\widetilde{w}_{k}^{T}\widetilde{x}_{ki}\right|\geq\epsilon\\
\epsilon & \left|1-\widetilde{w}_{k}^{T}\widetilde{x}_{ki}\right|<\epsilon
\end{cases},\quad\forall k,\ \forall i\in C_{k}.
\end{equation}

The solution for the concatenated weight vector $\widetilde{w}$ merely requires some turgid algebra and is relegated to Appendix~\ref{app_for_linear}. The pseudo code of the algorithm is presented in Algorithm~\ref{alg:algorithm_1}. We briefly mention here that the regularization coefficients for soft constraints on weight vectors $\left(\alpha\right)$ and hinge loss terms $\left(\beta\right)$ should be chosen in such a way that the Hessian obtained after concatenating the weight vectors is positive definite (as shown in Appendix~\ref{app_for_linear}).
\begin{algorithm}
\begin{algorithmic}[1]
\State{\textbf{Input:}\ $\widetilde{x}_{ki}, \widetilde{D}$ (diagonal matrix with $D$), $\widetilde{D}_{0}$ (off-diagonal matrix with $D$)}
\State{\textbf{Initialize} $z_{ki} = 1 \quad \forall \, k, \ \forall \, i \in C_k$}
\While{not converged} 
\State{Compute $\widetilde{X}$ as a diagonal matrix with entries $\sum_{i\in C_{k}}\frac{1}{4z_{ki}}\widetilde{x}_{ki}\widetilde{x}_{ki}^{T}$}
\State{Solve $\widetilde{w}= \frac{\beta}{2}\left(\widetilde{D}+\frac{\alpha}{2}\widetilde{D}_{0}+\beta\widetilde{X}\right)^{-1}\widetilde{x}$\ (KKT conditions) where {$\widetilde{w}=\left[\widetilde{w}_{1}^{T},\widetilde{w}_{2}^{T},\ldots,\widetilde{w}_{K}^{T}\right]^{T}$}} 
\State{Compute $z_{ki}=\left|1-\widetilde{w}_{k}^{T}\widetilde{x}_{ki}\right| \quad \forall \, k, \ \forall \, i \in C_k$} 
\EndWhile 
\State{\textbf{Output:}\ $\widetilde{W}=\left\{\widetilde{w}_{1},\widetilde{w}_{2},\ldots,\widetilde{w}_{K}\right\}$ }
\end{algorithmic}
\caption{Quadratic majorizer for multiclass/multilabel SVM with soft constraint on  weight vectors $W$  and hard constraint on biases $b$}
\label{alg:algorithm_1}
\end{algorithm}
\subsection{Incorporating soft and hard constraints into the formulation}\label{linear soft and hard constraints}
The formulation in the previous section imposed a soft constraint on the hyperplane normal vectors, wherein the inner product of every pair of vectors was minimized. Furthermore, a hard constraint on the biases $b_{k}$ was used to make their sum equal to zero. However, depending on the problem and dataset, we could also consider hard constraints on both the normal vectors 
$\left(\sum_{k=1}^{K}w_{k}=0\right)$ and the biases $\left(\sum_{k=1}^{K}b_{k}=0\right)$. Table~\ref{tab:linear_formulation_terms} provides the four different possibilities for incorporating the constraints into the formulation. For the sake of simplicity, we omit the margin terms and hinge loss terms which are common to all cases. In all these models, hyperparameters should be chosen in such a way that the Hessian of the objective function remains positive definite, ensuring that the optimization problem remains bounded from below. The details of the formulation specific to the Hessian for the case of soft $w$-hard $b$ are relegated to Appendix~\ref{app_for_linear}. 
\begin{table}[!htbp]
\centering
\begin{tabular}{|c|l|}
\hline
Constraints       & Appearance in objective (Lagrangian) function\\ \hline
soft $w$-soft $b$ & $\alpha\sum_{k=1}^{K}\sum_{l=k+1}^{K}w_{k}^{T}w_{l}+\;\gamma\left(\sum_{k=1}^{K}b_{k}\right)^{2}$                          \\ \hline
soft $w$-hard $b$ & $\alpha\sum_{k=1}^{K}\sum_{l=k+1}^{K}w_{k}^{T}w_{l}+\;\lambda\sum_{k=1}^{K}b_{k}$                                 \\ \hline
hard $w$-soft $b$ &  $\lambda\sum_{k=1}^{K}w_{k}\;+\gamma\left(\sum_{k=1}^{K}b_{k}\right)^{2}$                                \\ \hline
hard $w$-hard $b$ &  $\lambda\sum_{k=1}^{K}w_{k}\;+\eta\sum_{k=1}^{K}b_{k}$   \\ \hline
\end{tabular}
\caption{Terms in the objective function corresponding to all possible combinations of soft and hard constraints. The parameters $\lambda$ and $\eta$ denote the Lagrange multipliers corresponding to the hard constraints.}
\label{tab:linear_formulation_terms}
\end{table}
\subsection{Equivalence to the binary SVM}\label{binary_svm_equivalance}
\begin{figure}
\centering
\begin{subfigure}[b] {.45\textwidth}
  \centering
  \includegraphics[width=\textwidth]{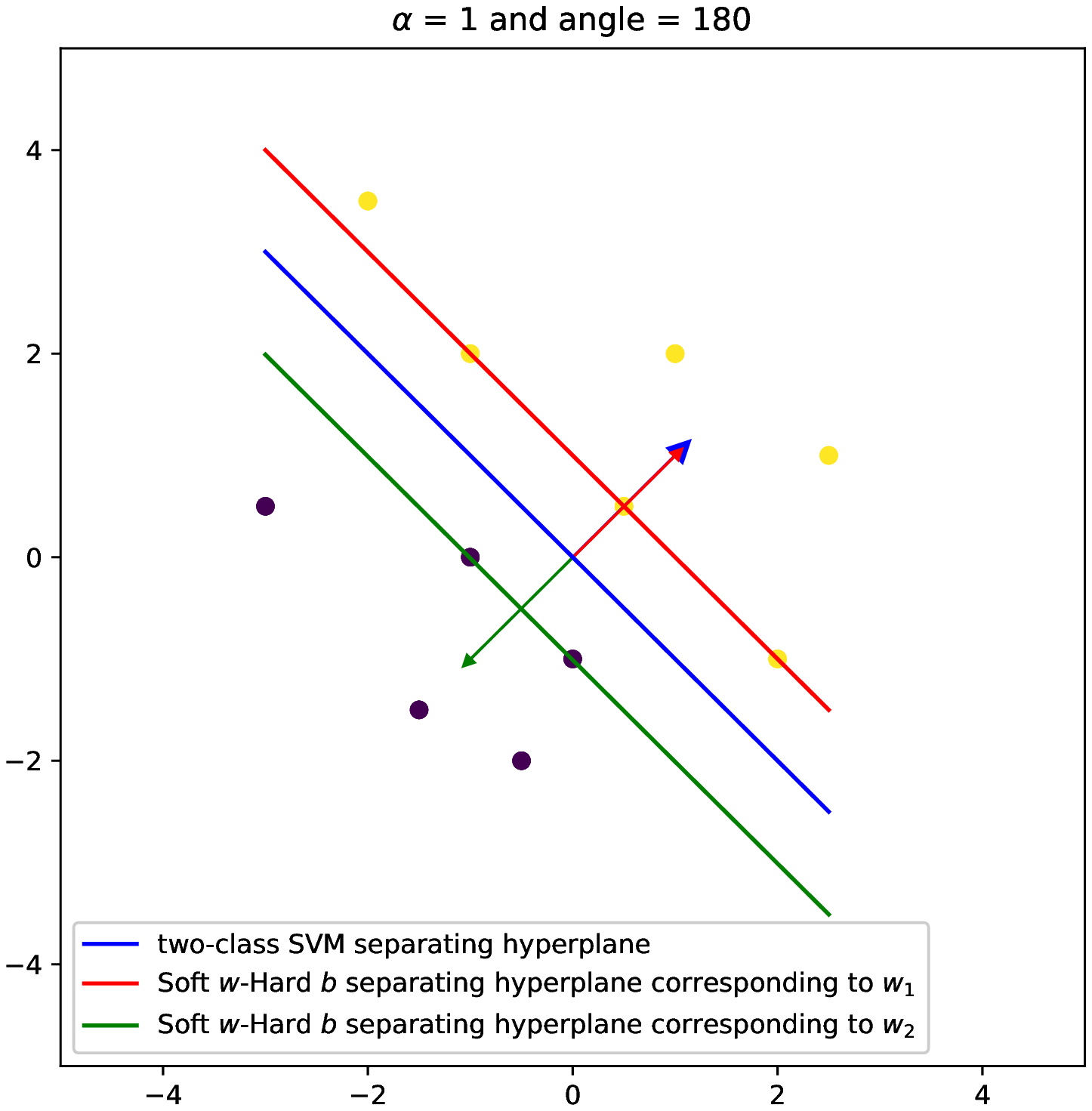}
  \caption{$\alpha=1$}
  \label{parallel-alpha_1}
\end{subfigure}
\hfill
\begin{subfigure}[b]{.45\textwidth}
  \centering
  \includegraphics[width=\textwidth]{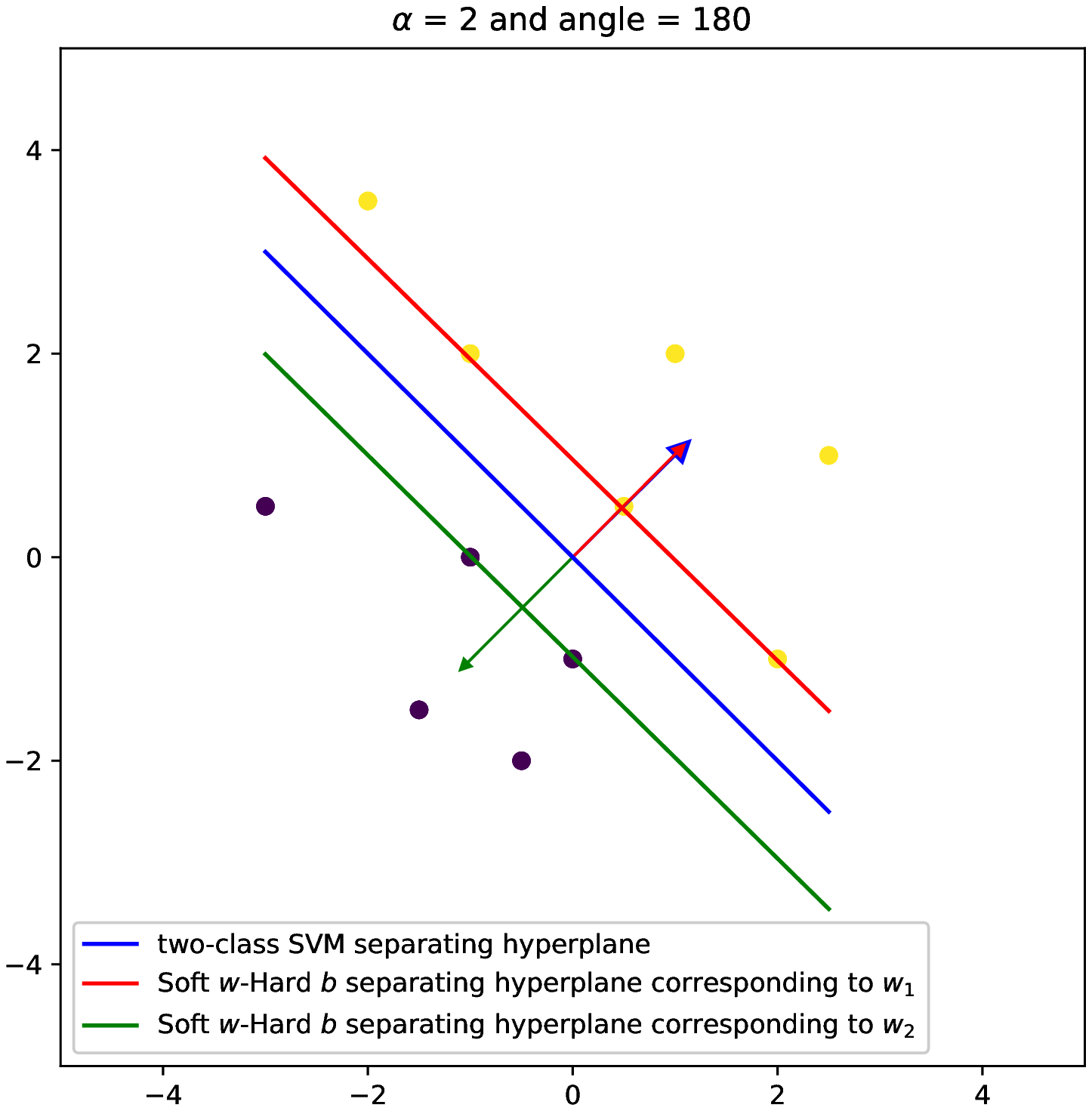}
  \caption{$\alpha=2$}
  \label{fig:sub2}
\end{subfigure}
\caption{Soft $w$-hard $b$ model results in parallel hyperplanes as the optimal solution. The angle between the normal vectors of the hyperplanes is $180$ degrees.}
\label{Sw-Hb parallel hyperplanes}
\end{figure}

The original two-class soft margin SVM problem is equivalent to the minimization of the following objective function:
\begin{equation}\label{equivalance}
    \begin{aligned}E(w,b)= & w^{T}w+C\sum_{i=1}^{N}[1-y_{i}(w^{T}x_{i}+b)]_{+}\\
= & w^{T}w+C\left[\sum_{i\in C_{1}}[1-(w^{T}x_{i}+b)]_{+}+\sum_{j\in C_{2}}[1-(-1)(w^{T}x_{j}+b)]_{+}\right]\\
= & w^{T}w+C\left[\sum_{i\in C_{1}}[1-(w^{T}x_{i}+b)]_{+}+\sum_{j\in C_{2}}[1-((-w)^{T}x_{j}+(-b)]_{+}\right].
\end{aligned}
\end{equation}
Note that we have used $w^{T}w$ instead of $\frac{1}{2}w^{T}w$ since the equivalence will be shown using two hyperplanes. From (\ref{equivalance}), we see that the existence of two hyperplanes with equal norms and anti-parallel normal vectors is implicit within the standard soft margin binary SVM formulation. The original binary SVM is equivalent to our formulation by imposing hard constraints on $w$ and $b$:
\begin{equation*}
    \begin{aligned}\min_{w_{1},w_{2},b_{1},b_{2}} & \begin{aligned}\quad & \frac{1}{2}w_{1}^{T}w_{1}+\frac{1}{2}w_{2}^{T}w_{2}+C\left[\sum_{i\in C_{1}}[1-(w_{1}^{T}x_{i}+b_{1})]_{+}+\sum_{j\in C_{2}}[1-(w_{2}^{T}x_{j}+b_{2})]_{+}\right]\end{aligned}\\
\textup{s.t.}\quad & \quad w_{1}+w_{2}=0,~\mathrm{and}\\
 & \quad b_{1}+b_{2}=0.
\end{aligned}
\end{equation*}
Incorporating these constraints into the objective function, we get 
\[
    E(w_{1},b_{1})=w_{1}^{T}w_{1}+C\left[\sum_{i\in C_{1}}[1-(w_{1}^{T}x_{i}+b_{1})]_{+}+\sum_{j\in C_{2}}[1+(w_{1}^{T}x_{j}+b_{1})]_{+}\right]
\]
which is identical to (\ref{equivalance}).
We have shown that when hard constraints are imposed on two weight vectors (making them anti-parallel but with equal norm) and on the two biases (making them sum to zero), we recover the original soft margin SVM. Furthermore, our soft $w$-hard $b$ model (shown in Table~\ref{tab:linear_formulation_terms})---while providing a degree of flexibility to the weight vectors' orientations and norms---is still capable of producing parallel hyperplanes (anti-parallel weight vectors) when the optimal solution dictates this to be the case. Consider the following example which is designed in such a way that the optimal solution must be two parallel hyperplanes. As shown in Figure~\ref{Sw-Hb parallel hyperplanes}, for any $\alpha\in\left[1,2\right]$, we get parallel hyperplanes and the solution coincides with the binary SVM.

\section{Generalization to RKHS}\label{Generalization to RKHS}
In Section~\ref{Unified SVM Classification Framework}, we presented the formulation and algorithm assuming linear separability between classes. However, linear separability is not a realistic assumption in many problems. Therefore, a more complex approach is required to extend these models to problems with nonlinear separability. In the following subsections, we give an introduction to kernels and demonstrate how the origin is formulated in this setting. We then develop the kernel formulation for unified multiclass and multilabel SVMs. 

\subsection{Kernels}\label{Kernels}
Kernels were introduced as a tool to map a nonlinearly separable dataset into an implicit higher (or infinite) dimensional reproducing kernel Hilbert space (RKHS) $\mathcal{H}$ where linear separability can be achieved without the need to explicitly compute the features in the transformed space \cite{vapnik2013nature}. In order to adapt our formulation to an RKHS, two problems have to be dealt with before we can proceed. As we will show, the origin of the RKHS needs to be properly treated and correctly interpreted in an infinite dimensional Hilbert space. Furthermore, the inclusion of a hard constraint on the weight vectors is not as straightforward as in the linearly separable case. We first briefly summarize how RKHS kernels are used in the SVM context.

Deploying a kernel function $\kappa:X\times X\rightarrow\mathbb{R}$, the inner product of feature mappings in a high dimensional RKHS is performed by function evaluation in the original space. If $\phi:X\rightarrow\mathbb{\mathcal{H}}$ is a mapping from the original space to a feature space, i.e., $X\ni x\mapsto\phi(x):=\kappa(\cdot,x)\in\mathbb{\mathcal{H}}$, then
\begin{equation*}
\left\langle\phi(x_{i}),\phi(x_{j})\right\rangle =\kappa(x_{i},x_{j}).    
\end{equation*}
By the representer theorem \cite{kimeldorf1971WahbaRepresenter}, each weight vector $w_{k}$ can be written as a linear combination of all projected patterns $\phi(x_{i}):=\kappa(\cdot,x_{i})$ in RKHS,
\begin{equation}\label{w_k-kernel}
w_{k}=\sum_{i=1}^{N}\alpha_{ki}\phi(x_{i}),\quad k\in\{1,\ldots,K\}.
\end{equation}

If we define $S_{W}$ as the subspace spanned by $W=\left\{ w_{1},w_{2},\cdots,w_{K}\right\}$, i.e., \begin{equation*}
S_{W}=\text{span}\left\{\phi(x_{1})\mathbin{,}\phi(x_{2})\mathbin{,}\cdots\mathbin{,}
\allowbreak \phi(x_{N})\right\},   
\end{equation*} 
we can write $x_{0}$ (the origin in the infinite dimensional feature space) as a summation of elements in $S_{W}$ and an element in the orthogonal complement of $S_{W}$, i.e., $S_{W}^{\perp}$
\begin{equation*}
x_{0}=\sum_{k=1}^{K}\mu_{k}w_{k}+s,\qquad s\in S_{W}^{\perp}.    
\end{equation*}
However, in calculating $\left\langle w_{k},x_{0} \right\rangle$, the inner product of all elements in $S_{W}^{\perp}$ taken with $w_{k}$ vanishes. Therefore, we consider only the $S_{W}$ component of $x_{0}$. Replacing $w_{k}$ in $x_{0}$ with (\ref{w_k-kernel}), we get
\begin{equation*}
    x_{0}=\sum_{k=1}^{K}\mu_{k}\sum_{i=1}^{N}\alpha_{ki}\phi(x_{i}).
\end{equation*}
In our application of RKHS principles, we need to compute the following inner products with respect to some kernel in RKHS:
\begin{equation*}
    \begin{aligned}\left\langle w_{k},w_{l}\right\rangle = & \;\left\langle \sum_{i=1}^{N}\alpha_{ki}\phi(x_{i}),\sum_{j=1}^{N}\alpha_{lj}\phi(x_{j})\right\rangle \\
= & \;\sum_{i=1}^{N}\sum_{j=1}^{N}\alpha_{ki}\alpha_{lj}\left\langle \phi(x_{i}),\phi(x_{j})\right\rangle \\
= & \;\sum_{i=1}^{N}\sum_{j=1}^{N}\alpha_{ki}\alpha_{lj}\kappa(x_{i},x_{j})\\
= & \;\alpha_{k}^{T}G\alpha_{l},
\end{aligned}
\end{equation*}
where $G$ is the Gram matrix formed by the set of pairwise RKHS inner products of patterns. Also, we have
\begin{equation*}
    \begin{aligned}\left\langle w_{k},\phi(x_{i})\right\rangle = & \;\left\langle \sum_{j=1}^{N}\alpha_{kj}\phi(x_{j}),\phi(x_{i})\right\rangle \\
= & \;\sum_{j=1}^{N}\alpha_{kj}\kappa(x_{j},x_{i})\\
= & \;\alpha_{k}^{T}g_{i},
\end{aligned}
\end{equation*}
where $g_{i}$ is the $i^\textup{th}$ column of the Gram matrix $G$.
\subsection{A Kernel Formulation for  Multiclass and Multilabel SVMs}
We now extend our formulation to an RKHS. The RKHS objective function directly follows (\ref{Lagrangian function for linear case}) with the main difference being the use of RKHS inner products:
\begin{equation*}
\begin{aligned}\mathcal{L}_{\mathrm{ker}}(W,x_{0})= & \sum_{k=1}^{K}\left\langle w_{k},w_{k}\right\rangle +\theta\sum_{k=1}^{K}\sum_{l=k+1}^{K}\left\langle w_{k},w_{l}\right\rangle \\
& +\beta\sum_{k=1}^{K}\sum_{i\in C_{k}}\left[1-\left\langle w_{k},\phi(x_{ki})-x_{0}\right\rangle \right]_{+}+\lambda\sum_{k=1}^{K}-\left\langle w_{k},x_{0}\right\rangle.
\end{aligned}
\end{equation*}
Again, after majorizing the hinge loss followed by a change of variables $b_{k}\equiv-\left\langle w_{k},x_{0}\right\rangle$, we have
\begin{equation*}
\begin{aligned}\mathcal{L}_{\mathrm{ker}}(W,b,\lambda)= & \sum_{k=1}^{K}\left\langle w_{k},w_{k}\right\rangle +\theta\sum_{k=1}^{K}\sum_{l=k+1}^{K}\left\langle w_{k},w_{l}\right\rangle \\
& +\beta\sum_{k=1}^{K}\sum_{i\in C_{k}}\frac{1}{4z_{ki}}\left[1+z_{ki}-\left(\left\langle w_{k},\phi(x_{ki})\right\rangle +b_{k}\right)\right]^{2}+\lambda \sum_{k=1}^{K}b_{k}.
\end{aligned}
\end{equation*}
Rewriting the Lagrangian using a matrix formulation of RKHS inner products, we get
\begin{equation*}
\begin{aligned}\mathcal{L}_{\mathrm{ker}}(\mathcal{A},b,\lambda,Z)= & \sum_{k=1}^{K}\alpha_{k}^{T}G\alpha_{k}+\theta\sum_{k=1}^{K}\sum_{l=k+1}^{K}\alpha_{k}^{T}G\alpha_{l}\\
& +\beta\sum_{k=1}^{K}\sum_{i\in C_{k}}\frac{1}{4z_{ki}}\left[1+z_{ki}-(\alpha_{k}^{T}g_{ki}+b_{k})\right]^{2}+\lambda\sum_{k=1}^{K}b_{k},
\end{aligned}
\end{equation*}
where $\mathcal{A} = \left\{\alpha_{1}, \alpha_{2},..., \alpha_{K}\right\}$. Again by concatenating $\alpha_{k}$ and $b_{k}$ into a vector $\widetilde{\alpha}_{k}$ and defining the following matrices
\begin{equation*}
\widetilde{\alpha}_{k}\equiv\left[\begin{array}{c}
\alpha_{k}\\
b_{k}
\end{array}\right],
\quad\widetilde{g}_{ki}\equiv\left[\begin{array}{c}
g_{ki}\\
1
\end{array}\right],
\quad G_{0}\equiv\left[\begin{array}{cc}
G & \overrightarrow{0}\\
\overrightarrow{0}^{T} & 0
\end{array}\right],
\quad e\equiv\left[\begin{array}{c}
0\\
\vdots\\
0\\
1
\end{array}\right]_{(N+1)},
\end{equation*}
the Lagrangian is rewritten as 
\[
\begin{aligned}\mathcal{L}_{\text{ker}}(\widetilde{\mathcal{A}},\lambda,Z)= & \sum_{k=1}^{K}\widetilde{\alpha}_{k}^{T}G_{0}\widetilde{\alpha}_{k}+\theta\sum_{k=1}^{K}\sum_{l=k+1}^{K}\widetilde{\alpha}_{k}^{T}G_{0}\widetilde{\alpha}_{l}\\
 & +\beta\sum_{k=1}^{K}\sum_{i\in C_{k}}\frac{1}{4z_{ki}}\left[1+z_{ki}-(\widetilde{\alpha}_{k}^{T}\widetilde{g}_{ki})\right]^{2}+\lambda\sum_{k=1}^{K}e^{T}\widetilde{\alpha}_{k},
\end{aligned}
\]
where $\widetilde{\mathcal{A}} = \left\{\widetilde{\alpha}_{1}, \widetilde{\alpha}_{2},..., \widetilde{\alpha}_{K}\right\}$. 
\subsection{Hard constraints on weight vectors and biases}
In a manner similar to the linear setting in Section~\ref{linear soft and hard constraints}, we can also incorporate any combination of soft and hard constraints on the normal vectors and biases into the formulation in the kernel setting. As mentioned above, using the representer theorem we can write each $w_{k}$ in an RKHS as a linear combination of mapped patterns. Note that we cannot directly incorporate a hard constraint on the weight vectors into the formulation as we did in the linear case since $0_{H}$ here is the origin of an infinite dimensional Hilbert space and is therefore not computable. One way to set the sum of the $w_{k}$ equal to $0_{H}$ in RKHS, i.e., $\sum_{k=1}^{K}w_{k}=0_{H}$, is by adding the following constraints to the problem:
\begin{equation}\label{sumo_of_alpha_k_zero}
\sum_{k=1}^{K}\alpha_{ki}=0, \quad i\in\{1,2,\ldots,N\}.    
\end{equation}
These constraints are derived from the fact that each $w_{k}$ is a linear combination of mapped patterns
\[
w_{k}=\sum_{i=1}^{N}\alpha_{ki}\;\phi(x_{i}), \quad k\in\{1,\ldots,K\}
\]
and therefore setting the sum to $0_{H}$ via
\[
\sum_{k=1}^{K}w_{k}=\sum_{k=1}^{K}\sum_{i=1}^{N}\alpha_{ki}\;\phi(x_{i})=\sum_{i=1}^{N}\sum_{k=1}^{K}\alpha_{ki}\;\phi(x_{i})=\sum_{i=1}^{N}\phi(x_{i})\sum_{k=1}^{K}\alpha_{ki}=0_{H}
\]
is equivalent to (\ref{sumo_of_alpha_k_zero}).
Here, we only present the formulation for the hard constraints on both weight vectors and biases. The detailed formulation for the case of soft $w$-hard $b$ is also presented in Appendix~\ref{app_kernel}. The other two cases can be obtained in similar fashion. The optimization problem is
\begin{equation} 
    \begin{aligned}\min_{\mathcal{A},b,Z}\quad & \frac{1}{2}\sum_{k=1}^{K}\alpha_{k}^{T}G\alpha_{k}+\beta\sum_{k=1}^{K}\sum_{i\in C_{k}}\frac{1}{4z_{ki}}\left[1+z_{ki}-(\alpha_{k}^{T}g_{ki}+b_{k})\right]^{2}\\
\textup{s.t.}\quad & \sum_{k=1}^{K}\alpha_{ki}=0,\quad i\in\{1,2,\ldots,N\}\\
 & \sum_{k=1}^{K}b_{k}=0.
\end{aligned}
\end{equation}
Similarly, by concatenating $b_{k}$ to $\alpha_{k}$, we get the simplified optimization problem
\begin{equation}
    \begin{aligned}\min_{\widetilde{\mathcal{A}},Z} & \quad\frac{1}{2}\sum_{k=1}^{K}\widetilde{\alpha}_{k}^{T}G_{0}\widetilde{\alpha}_{k}+\beta\sum_{k=1}^{K}\sum_{i\in C_{k}}\frac{1}{4z_{ki}}\left[1+z_{ki}-(\widetilde{\alpha}_{k}^{T}\widetilde{g}_{ki})\right]^{2}\\
\textup{s.t.} & \quad\sum_{k=1}^{K}\widetilde{\alpha}_{k}=0.
\end{aligned}
\end{equation}
We then concatenate all $\widetilde{\alpha}_{k}$ into one vector, $\widetilde{\alpha}$, and solve for it using KKT optimality conditions. The four cases for the kernel formulation with different settings of constraints are summarized in Table~\ref{tab:kernel_formulation_terms}. Similar to the linear case in Section~\ref{Unified SVM Classification Framework}, regularization terms and hinge loss terms are left out since they are common to all cases.

\begin{table}
\centering
\begin{tabular}{|c|l|}
\hline
Constraints       & Appearance in objective (Lagrangian) function \\ \hline
soft $w$-soft $b$ & $\theta\sum_{k=1}^{K}\sum_{l=k+1}^{K}\alpha_{k}^{T}G\alpha_{l}+\gamma\left(\sum_{k=1}^{K}b_{k}\right)^{2}$                           \\ \hline
soft $w$-hard $b$ & $\theta\sum_{k=1}^{K}\sum_{l=k+1}^{K}\alpha_{k}^{T}G\alpha_{l}+\eta\sum_{k=1}^{K}b_{k}$                                 \\ \hline
hard $w$-soft $b$ &  $\sum_{i=1}^{N}\lambda_{i}\sum_{k=1}^{K}\alpha_{ki}+\gamma\left(\sum_{k=1}^{K}b_{k}\right)^{2}$                                \\ \hline
hard $w$-hard $b$ &  $\sum_{i=1}^{N}\lambda_{i}\sum_{k=1}^{K}\alpha_{ki}+\eta\sum_{k=1}^{K}b_{k}$                                \\ \hline
\end{tabular}
\caption{Objective functions corresponding to all combinations of soft and hard constraints in RKHS. The parameters $\left\{\lambda_{i}\right\}$ and $\eta$ denote the Lagrange multipliers for hard constraints}
\label{tab:kernel_formulation_terms}
\end{table}

\section{Experiments}\label{experiments}
In this section, we present our results for some well known datasets using our models and compare them to the OvR approach with the soft margin two-class SVM as the base classifier---using the scikit-learn \cite{pedregosa2011scikit} implementation. In Section~\ref{Generating unseen label set configuration}, we discuss one of the drawbacks of the OvR approach for a toy dataset which does not have samples exclusively drawn from each of the classes. In Section~\ref{multiclass datasets}, we first present the results for a few two-class datasets and compare the performance of our model with the soft margin binary SVM. Subsequently, we present the results of our model on some famous multiclass datasets and compare them with OvR and the Crammer-Singer (CS) model for the linear case. In Section~\ref{multilabel datasets}, our model is tested on benchmark multilabel datasets and the results are compared to binary relevance (BR).
\begin{figure}[!tbp]
\centering
  \includegraphics[trim=1cm 1cm 1cm 2cm, clip=true, width=0.4\textwidth]{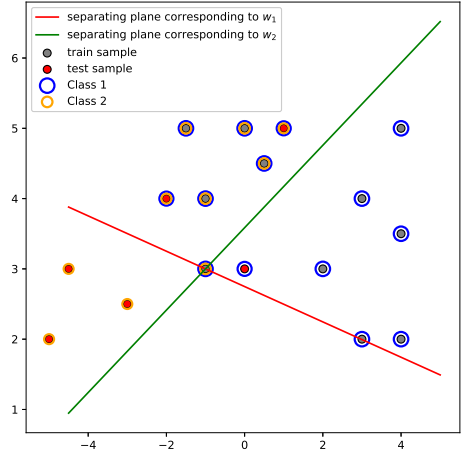}
\caption{\text{OvN} determines the boundary for patterns in a class without requiring exclusive oppositional patterns from other classes.}
\label{unseen_label}
\end{figure}
We use 3-fold cross validation to choose the hyperparameters. We assume a finite and discrete grid of parameter values. First, the data is shuffled and divided into three equal folds. We choose the first tuple of grid points and train the model on two folds (training set). Then the trained model with this tuple is validated on the $3^{\textup{rd}}$ fold (validation set) and the model accuracy on the validation set is recorded. 
We repeat this procedure three times, choosing a different fold as validation set on each occasion and the remaining two folds as the training set and record all accuracy measures. In this way, we get three accuracy measures for the first choice of hyperparameters. This procedure is repeated for each and every tuple of grid points. Once again, all accuracy measures are recorded and the average accuracy on three folds for each tuple is computed. The maximum average accuracy is reported here. We tested our models on known datasets and compare the results against frequently used classification models in the literature.
\subsection{Generating unseen label set configuration}\label{Generating unseen label set configuration}
One of the drawbacks of the BR approach for multilabel problems is that we might not have data only belonging to one class in order to train its classifier. For illustration, consider the following scenario shown in Figure~\ref{unseen_label}. In this two-class example, the training set patterns belong either to class 1 (label $\left[1,0\right]$) or to both classes (label $\left[1,1\right]$). Therefore there is no pattern which only belongs to class 2. Obviously, the \text{OvR} approach can not train a classifier for class 1 versus class 2 as there is no pattern belonging solely to the latter. Hence we need an approach for determining the hyperplanes of each class \emph{without requiring data belonging to opposing classes}. In other words, we need each class to determine its classifier based on available information of its members without the need for direct opposition from other classes. This is the core idea of our approach and hence the monicker one versus none (OvN).

Scikit-learn handles this situation by just assigning the label $1$ to any test set pattern and therefore only produces label set $\left[1,0\right]$ or $\left[1,1\right]$ for each training or test pattern. The cross training approach of \cite{boutell2004SceneSVM} is also not able to cope with this situation. They implemented the OvR approach to train separate classifiers. If a pattern belongs to both class 1 and class 2, it is considered a class 1 (class 2) pattern while training the class 1 (class 2) classifier. However, even this strategy is not able to mitigate the situation. With the OvN approach, we are capable of producing new label sets which have not occurred during training. Figure~\ref{unseen_label} demonstrates this scenario. As we can see from Figure~\ref{unseen_label}, it does not make sense to assign label set $\left[1,1\right]$ to patterns $\left(-4,3\right)$, $\left(-5,2\right)$ and $\left(-3,3\right)$ since they are far away from the points which have both labels. Our linear soft $w$-hard $b$ model, assigns label set $\left[0,1\right]$ to these points. 
\begin{table}
\centering
\begin{tabular}{|c|c|c|c|}
\hline
test pattern & ground truth & OvN (soft $w$-hard $b$) & OvR (scikit-learn) \\ \hline
$\left(-4,3\right)$ &  $\left[0,1\right]$  &   $\left[0,1\right]$  &   $\left[1,1\right]\times$  \\ \hline
$\left(-5,2\right)$ &  $\left[0,1\right]$  &   $\left[0,1\right]$  &   $\left[1,1\right]\times $ \\ \hline
$\left(-3,3\right)$ &  $\left[0,1\right]$  &   $\left[0,1\right]$  &   $\left[1,1\right]\times $  \\ \hline
$\left(-2,4\right)$ &       $\left[1,1\right]$    &    $\left[1,1\right]$  &   $\left[1,1\right]$  \\ \hline
$\left(0,3\right)$  &       $\left[1,0\right]$    &   $\left[1,0\right]$   &   $\left[1,0\right]$  \\ \hline
$\left(1,5\right)$  &       $\left[1,1\right]$    &   $\left[1,1\right]$   &   $\left[1,1\right]$  \\ \hline
\end{tabular}
\caption{OvR approach fails to obtain a classifier for patterns which only belong to class 2 since there are no such patterns in the training set. Scikit-learn assigns all test patterns to class 1 ($\times$ indicates wrongly assigned labels). OvN can assign correct label ($\left[0,1\right]$) to these patterns.}
\label{kernel_objective_terms}
\end{table}
\subsection{Multiclass datasets}\label{multiclass datasets}
\begin{figure}
  \centering
  \subcaptionbox{\label{Hourglass with Gaussian kernel}}{\includegraphics[trim=1.5cm 1.2cm 1.5cm 2cm, clip=true, width=2in]{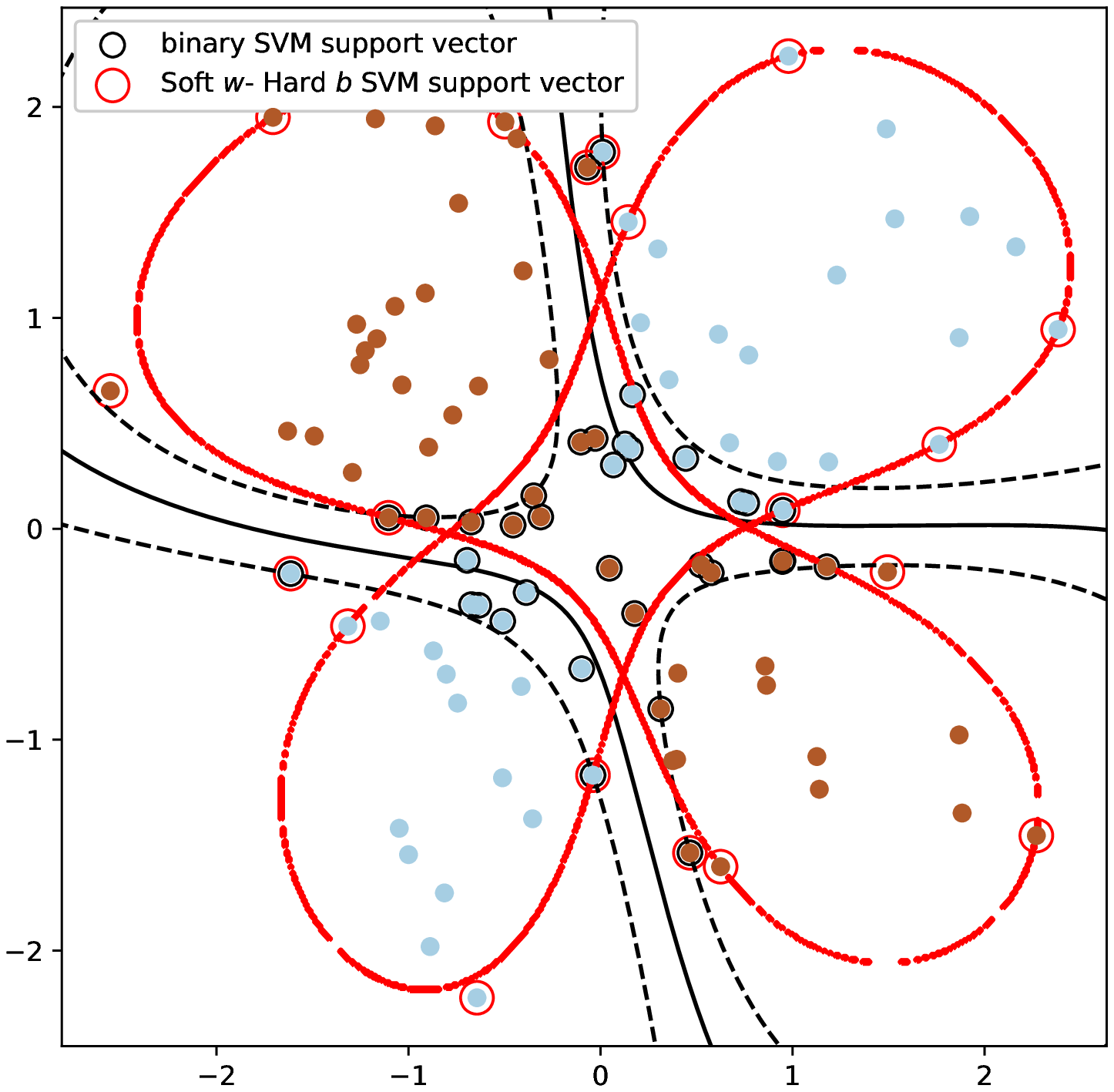}}\hspace{1em}%
  \subcaptionbox{\label{Hourglass with polynomial kernel}}{\includegraphics[trim=1.5cm 1.2cm 1.5cm 2cm, clip=true, width=2in]{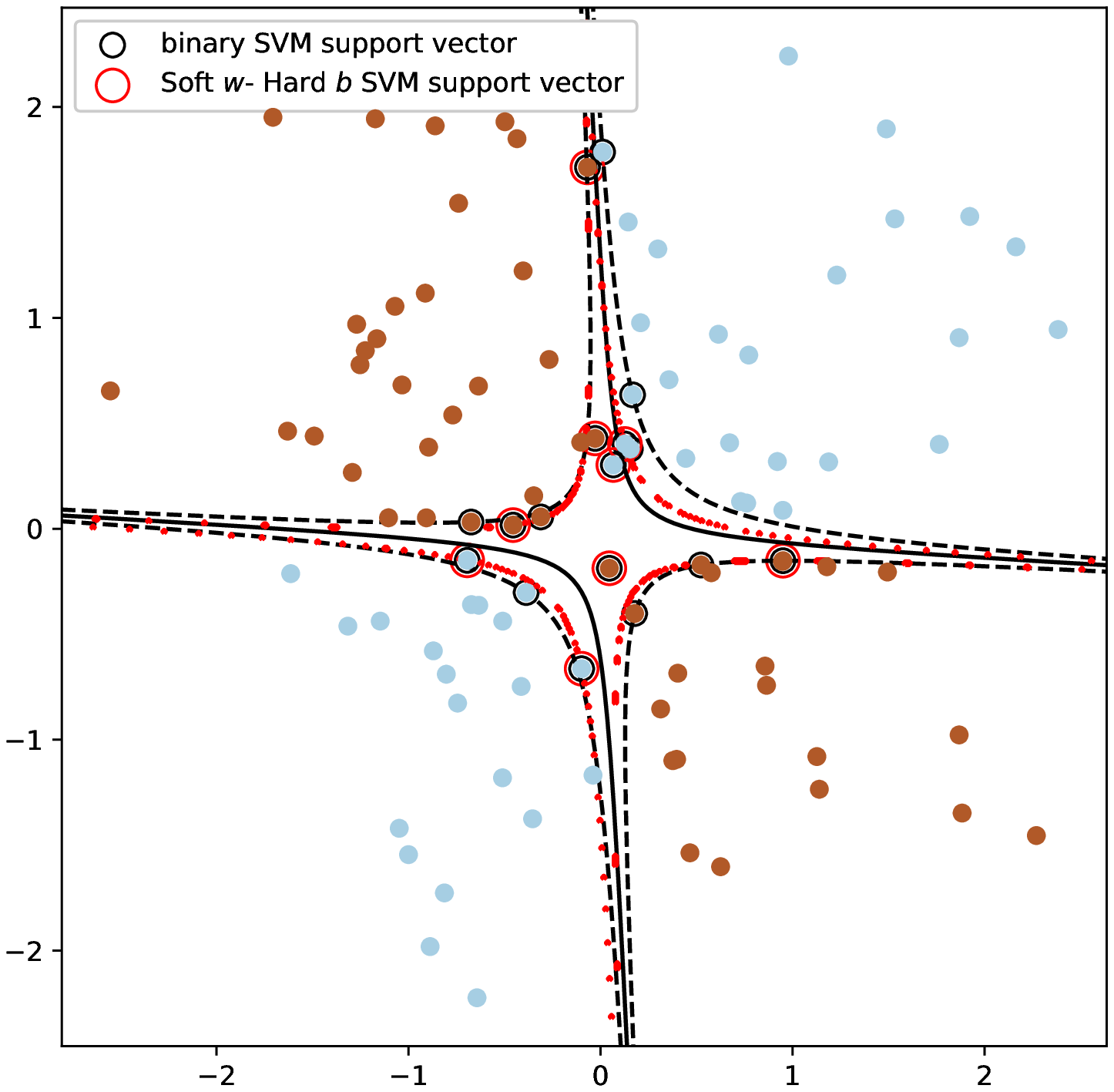}}
  \subcaptionbox{\label{Moon with Gaussian kernel}}{\includegraphics[trim=1.5cm 1.2cm 1.5cm 1.5cm, clip=true, width=2in]{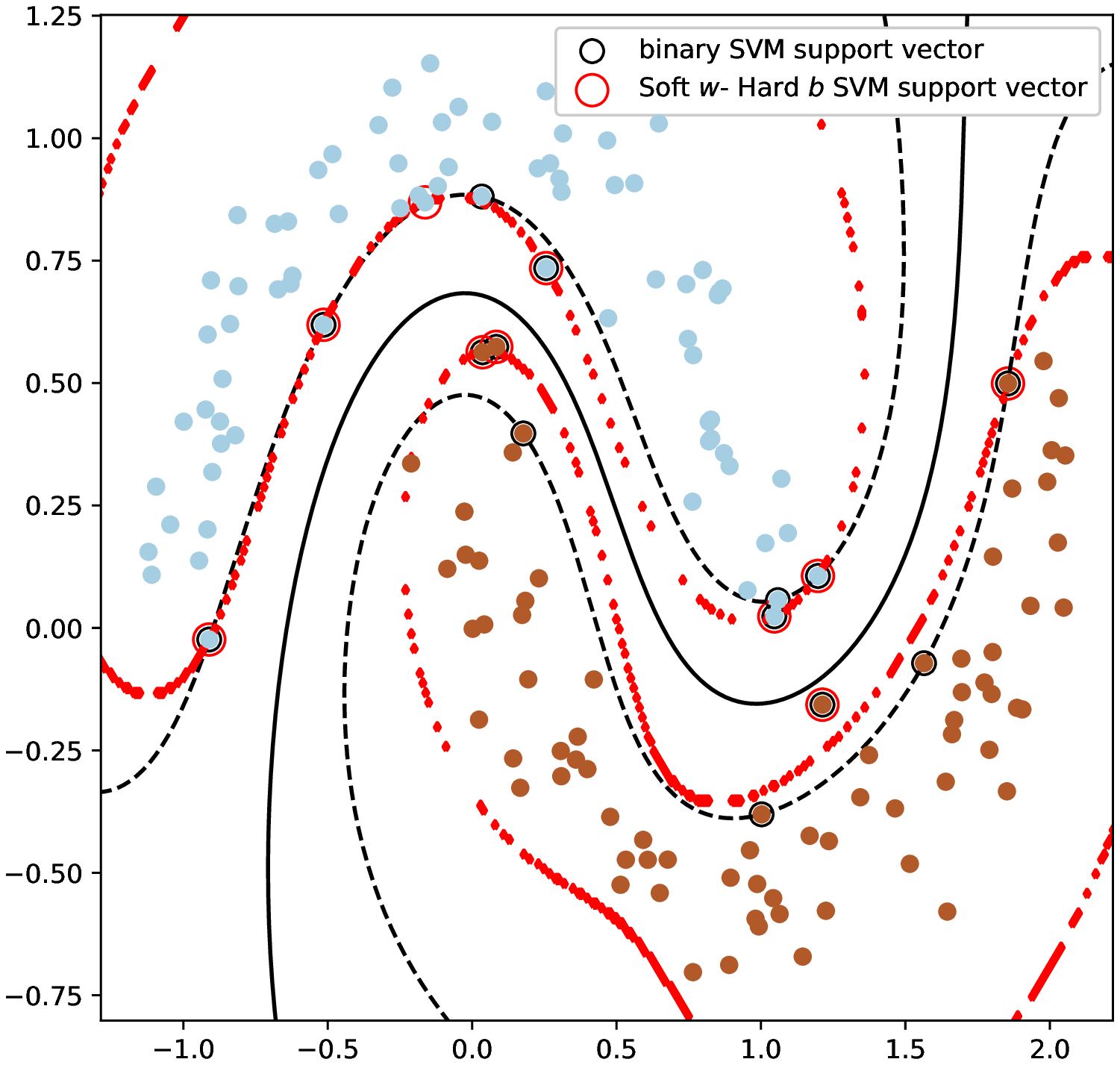}}\hspace{1em}%
  \subcaptionbox{\label{Two-class with linear kernel}}{\includegraphics[trim=1.5cm 1.2cm 1.5cm 1.5cm, clip=true, width=2in]{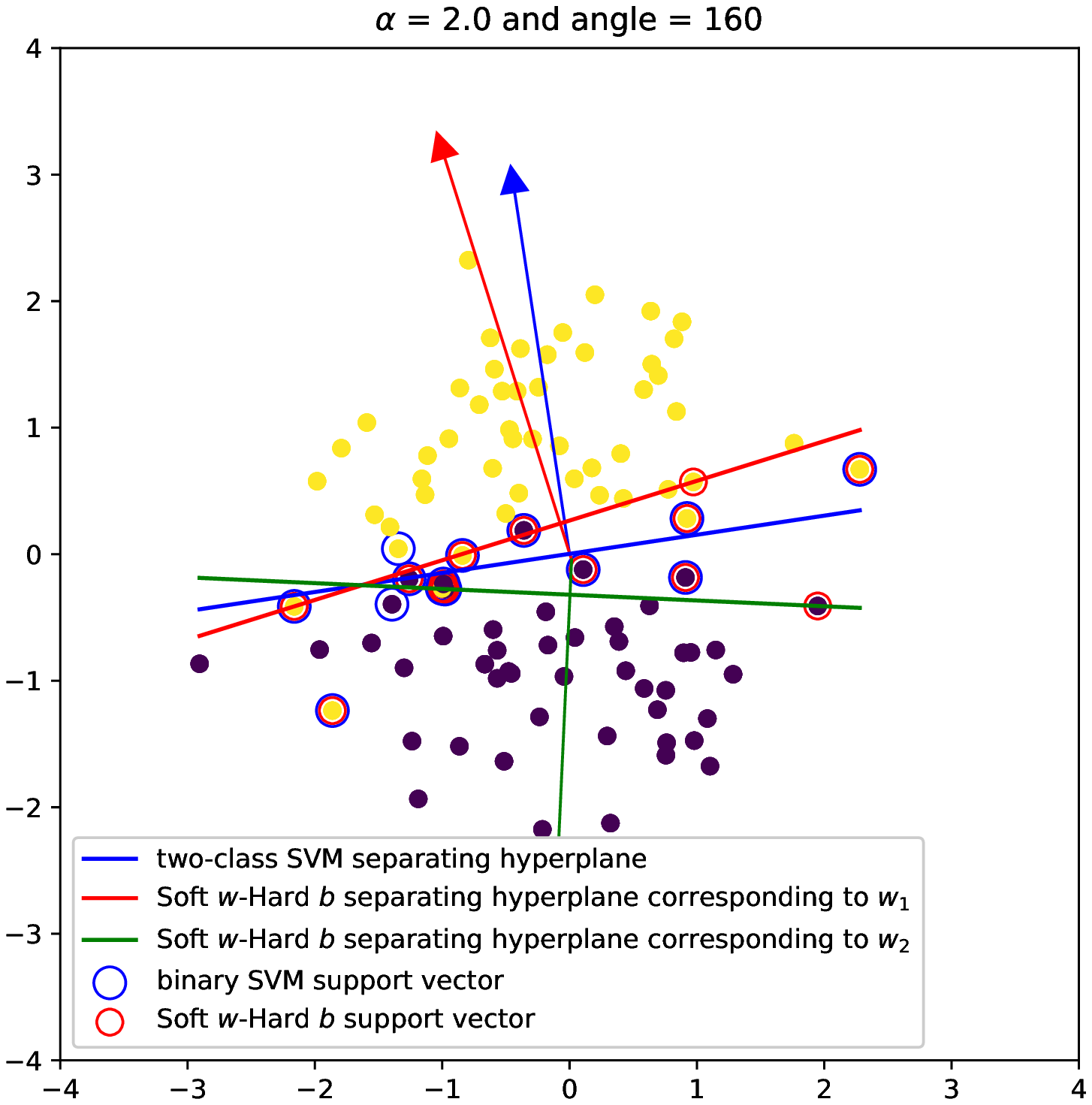}}\hspace{1em}%
\caption{Applying the soft $w$-hard $b$ model and binary SVM with linear and nonlinear kernels on different datasets. (\subref{Hourglass with Gaussian kernel}) Hourglass with Gaussian kernel; (\subref{Hourglass with polynomial kernel}) Hourglass with polynomial kernel; (\subref{Moon with Gaussian kernel}) Moon with Gaussian kernel; (\subref{Two-class with linear kernel}) Two classes with linear kernel. Patterns with projection less than 1 on both weight vectors, are assigned to the class with greater projection magnitude.} 
\label{4_images}
\end{figure}

We first show the results of applying our model (with linear and nonlinear kernels) to 2 class problems for 2-dimensional datasets for illustrative purposes. The column SVM in Table~\ref{performance_comparison_two_class} is the original two-class SVM implemented by scikit-learn. The metric for evaluating multiclass learning algorithms is the accuracy score. We also demonstrate the choice and number of support vectors for each model. In this section, we mainly use the Gaussian Radial Basis Function (GRBF) kernel which for two patterns $x_{i}$ and $x_{j}$ is defined as $\kappa\left(x_{i},x_{j}\right)=\text{exp}\left(-\frac{1}{2\sigma^{2}}\|x_{i}-x_{j}\|_{2}^{2}\right)$ where $\sigma$ is a hyperparameter.
\begin{table}
\centering
\begin{tabular}{|c|c|c|c|c|c|}
\hline
\multicolumn{1}{|c|}{\multirow{2}{*}{Data}} & \multirow{2}{*}{Kernel}       & \multicolumn{2}{c|}{Accuracy}                 & \multicolumn{2}{c|}{No. of sv}                \\ \cline{3-6} 
\multicolumn{1}{|c|}{}                      &                               & S$w$-H$b$                  & SVM                   & S$w$-H$b$                 & SVM                   \\ \hline
\multirow{2}{*}{hourglass}                  & \multicolumn{1}{l|}{Gaussian} & \multicolumn{1}{c|}{1} & \multicolumn{1}{c|}{0.97} &  \multicolumn{1}{c|}{$\left[9,10\right]$} & \multicolumn{1}{c|}{$\left[17,18\right]$} \\ \cline{2-6} 
                                            & Poly(deg.2)                   & 1                     & 0.99                       &  $\left[5,4\right]$                     &   $\left[8,9\right]$                    \\ \hline
random                                      & Linear                        & 0.95                     & 0.87                      &  $\left[6,6\right]$                    &   $\left[7,7\right]$                    \\ \hline
moon                                        & Gaussian                      & 1                     & 1                      & $\left[4,6\right]$                      &  $\left[7,7\right]$                     \\ \hline
\end{tabular}
\caption{Performance comparison of our model with binary SVM. OvN soft $w$-hard $b$ (S$w$-H$b$) results in better accuracy with fewer support vectors}
\label{performance_comparison_two_class}
\end{table}
The hyperplanes corresponding to the classifiers in each model as well as the support vectors are depicted in Figure~\ref{4_images}.

Table~\ref{tab:Results for multiclass} shows the results of our model, OvN, compared to OvR with binary SVM with linear and nonlinear kernels as well as the Crammer-Singer (CS) model---using the implementation in scikit-learn \cite{pedregosa2011scikit}. Three multiclass (with more than two classes) benchmark datasets are chosen. We normalized the datasets and did not use any supervised or unsupervised dimensionality reduction. In Table~\ref{tab:Results for multiclass}, the average accuracy and loss on left out folds in 3-fold cross validation are reported.
\begin{table}[!htbp]
\centering
\begin{tabular}{|c|c|c|c|c|c|c|c|}
\hline
\multirow{2}{*}{Dataset} & \multirow{2}{*}{Kernel} & \multicolumn{2}{c|}{OvN S$w$-H$b$} & \multicolumn{2}{c|}{OvR (scikit-learn)} & \multicolumn{2}{l|}{CS (scikit-learn)} \\ \cline{3-8} 
                         &                         & AS               & HL              & AS                 & HL                 & AS                 & HL                \\ \hline
\multirow{2}{*}{glass}   & L                       & \textbf{0.851}           & 0.149           & 0.598              & 0.402              & 0.668              & 0.332             \\ \cline{2-8} 
                         & G                       & \textbf{0.869}            & 0.131           & 0.701              & 0.299              & -                  & -                 \\ \hline
\multirow{2}{*}{iris}    & L                       & \textbf{0.987}            & 0.013           & 0.960              & 0.04               & 0.967              & 0.033             \\ \cline{2-8} 
                         & G                       & \textbf{0.987}            & 0.013           & 0.967              & 0.033              & -                  & -                 \\ \hline
\multirow{2}{*}{wine}    & L                       & 0.843            & 0.157           & 0.967             & 0.033               & \textbf{0.983}              & 0.017             \\ \cline{2-8} 
                         & G                       & 0.966            & 0.034           & \textbf{0.986}              & 0.014              & -                  & -                 \\ \hline
\end{tabular}
\caption{Results for multiclass with average accuracy on 3-fold cross validation with linear (L) and Gaussian (G) kernels.}
\label{tab:Results for multiclass}
\end{table}
\subsection{Multilabel datasets}\label{multilabel datasets}
Multilabel classification is a generalization of the multiclass problem where more than one label is assigned to each pattern. There are many applications in which a multilabel framework is required to assign instances to more than one class if relevant. The scene dataset \cite{boutell2004SceneSVM} is one of the most widely used multilabel classification datasets. It has 294 features, 2407 instance images and 6 classes. We implemented our unified soft $w$-hard $b$ OvN model and compared the results to the OvR (BR) approach with the soft margin two-class SVM for multilabel classification---using the scikit-learn implementation. We also implemented our model on the emotions (music) dataset \cite{trohidis2008music}. It consists of 593 songs with 6 labels and 72 features.

There are different ways of evaluating the performance of a multilabel classifier since each label set prediction can be completely incorrect, completely correct or partially correct with different levels of correctness in the multilabel setting \cite{sorower2010literature}. Two main types of metrics are introduced in the literature: bipartition evaluation metrics and ranking evaluation metrics. The bipartition category in turn has two subcategories of instance-based (example-based) and label-based metrics. Our evaluation in this work is based on instance-based metrics. The Hamming loss \cite{schapire1999hammingloss}---an instance-based metric---is defined as the fraction of incorrectly predicted labels to the total number of labels averaged over all instances and normalized by the number of classes
\begin{equation*}
\text{Hamming loss}\equiv\frac{1}{NK}\sum_{n=1}^{N}\left|P_{n}\Delta Y_{n}\right|,
\end{equation*}
where $Y_{n}$ is the set of actual (true) labels, $P_{n}$ is the set of predicted labels and $\Delta$ denotes the symmetric difference of two sets. The ideal value of the Hamming loss is zero. Exact Match Ratio (subset accuracy) is a strict example-based metric and computes the ratio of instances whose set of predicted labels exactly matches their corresponding set of true labels, i.e., $\frac{1}{N}\sum_{n=1}^{N}I\left(P_{n}=Y_{n}\right)$ where $I\left(\cdot\right)$ is the indicator function.
Other instance-based metrics borrowed from information retrieval are accuracy, precision, recall and $F_{1}$-score which is the harmonic mean of precision and recall \cite{godbole2004discriminative}. These metrics account for partially correct predictions. In accuracy score, the proportion of correctly predicted labels to the total number of labels (predicted and true) for each instance is obtained and averaged over all instances \cite{gibaja2015tutorial}:
\begin{equation*}
    \text{Accuracy}=\frac{1}{N}\sum_{n=1}^{N}\frac{\left|P_{n}\cap Y_{n}\right|}{\left|P_{n}\cup Y_{n}\right|}.
\end{equation*}
In our experiments, we use accuracy and Hamming loss metrics for comparison.
Table~\ref{tab:multilabel-datasets} shows the results of implementing our model on a synthetic multilabel dataset as well as the scene and emotions dataset. As we can see, the performance is slightly better in the case of OvR. However, we need to keep two points in mind. First, our soft $w$-hard $b$ model is a unified one which attempts to learn all classifiers in one shot. So in this sense, OvR which separately learns the classifiers has a better chance of focusing on each class. However, it completely ignores the relationships between labels. Our model on the other hand considers the interrelationships between each pair of labels through soft constraints imposed on the weight vectors. Second, the label selection criteria are very strict in our model. We only assign a label to a pattern if its projection to the corresponding weight vector is greater than or equal to 1. Designing a less strict decision criterion which assigns labels to projections less than 1---provided it meets other conditions---may improve performance. This is an exciting avenue for future research.
\begin{table}[!htbp]
\centering
\begin{tabular}{|c|c|c|c|c|c|}
\hline
\multirow{2}{*}{Dataset}                         & \multirow{2}{*}{Kernel} & \multicolumn{2}{c|}{OvN S$w$-H$b$} & \multicolumn{2}{c|}{OvR (BR)} \\ \cline{3-6} 
                                                 &                         & AS               & HL              & AS            & HL            \\ \hline
\multicolumn{1}{|l|}{\multirow{2}{*}{synthetic}} & L                       & 0.837            & 0.133           & 0.828         & 0.14          \\ \cline{2-6} 
\multicolumn{1}{|l|}{}                           & G                       & 0.843            & 0.142           & 0.867         & 0.122         \\ \hline
\multirow{2}{*}{scene}                           & L                       & 0.557            & 0.177           & 0.601         & 0.10          \\ \cline{2-6} 
                                                 & G                       & 0.652            & 0.12            & 0.684         & 0.08          \\ \hline
\multirow{2}{*}{emotions}                        & L                       & 0.494            & 0.235           & 0.530         & 0.195         \\ \cline{2-6} 
                                                 & G                       & 0.50             & 0.255           & 0.575         & 0.175         \\ \hline
\end{tabular}
\caption{Results for multilabel datasets with average accuracy score (AS) and Hamming loss (HL) on all folds. Linear (L) and Gaussian (G) kernels are applied.}
\label{tab:multilabel-datasets}
\end{table}
\section{Conclusion}\label{Conclusion}
In this paper, we began by motivating the need for a new multiclass and multilabel SVM. Contrary to popular belief, a multilabel SVM is not a simple extension of previous SVMs and this impacts our multiclass SVM formulation as well---an unexpected development. At its core, the new formulation employs a separate hyperplane for each class and uses (soft or hard) constraints to incorporate class separation. From a technical perspective, the new approach results in class specific weight vectors and reformulation of the origin as biases in feature vector space. Extension to RKHS kernels requires even more careful treatment of the origin in Hilbert space. Hard and soft constraints on the weight vectors and the biases (related to the origin) can be combined and we showcased a few examples in both the multiclass and multilabel settings. The experiments clearly demonstrate a competitive classifier which eschews \emph{both} the one versus all and one versus one dichotomies widely prevalent at the present time.

The core idea---separate non anti-parallel hyperplanes for each class---can be extended to other classifiers as well such as the Fisher linear discriminant. This is a possible avenue for future research. Explicit inter label dependencies were not considered in the present work and could be incorporated in future work. We can also consider a formulation similar to the $\nu$-SVM in the future and incorporate margin controlling variables as well. In terms of optimization algorithms, we developed a very simple iterative majorization approach in the primal. A more elaborate subgradient-based approach can be considered in the future. Furthermore, we have not investigated the dual formulation in our setting. This appears to be a straightforward extension. Finally, better schemes for hyperparameter estimation via nested cross-validation and the like can be worked out. 

From a larger perspective, it is entirely straightforward to incorporate negative labels in our framework---theoretically speaking. The main reason why this was not developed in the present work is due to the severe lack of training sets with negative labels (such as ''I don't know what I am but I'm definitely NOT A libertarian'' in a political survey). If such information is available in future datasets, then we can flesh out the space of negative labels in our framework. This is an interesting avenue for future work.

\section*{Acknowledgement}{This work is partially supported by NSF IIS 1743050 to A.R.}

\newpage
\appendix
\section{Optimization scheme for linearly separable case with soft constraint on weight vectors and hard constraint on biases} \label{app_for_linear}
Rewriting the Lagrangian in (\ref{Lagrangian function linear case}) with the matrices introduced in (\ref{matrices soft w hard b}), we have
\begin{equation}\label{extended formulation linear case Lagrangian function}
    \begin{aligned} \mathcal{L}(\widetilde{W},\lambda,Z)= & \sum_{k=1}^{K}\widetilde{w}_{k}^{T}D\widetilde{w}_{k}+\alpha\sum_{k=1}^{K}\sum_{l=k+1}^{K}\widetilde{w}_{k}^{T}D\widetilde{w}_{l}\\
 & +\beta\sum_{k=1}^{K}\sum_{i\in C_{k}}\frac{1}{4z_{ki}}\left[1+z_{ki}-(\widetilde{w}_{k}^{T}\widetilde{x}_{ki})\right]^{2}+\lambda\sum_{k=1}^{K}e^{T}\widetilde{w}_{k}
\\
= & \sum_{k=1}^{K}\widetilde{w}_{k}^{T}D\widetilde{w}_{k}+\sum_{k=1}^{K}\sum_{l=k+1}^{K}\widetilde{w}_{k}^{T}D\widetilde{w}_{l}+ \lambda\sum_{k=1}^{K}e^{T}\widetilde{w}_{k}\\
 & +\beta\sum_{k=1}^{K}\sum_{i\in C_{k}}\frac{1}{4z_{ki}}\left[\widetilde{w}_{k}^{T}\widetilde{x}_{ki}\widetilde{x}_{ki}^{T}\widetilde{w}_{k}-2(1+z_{ki})(\widetilde{w}_{k}^{T}\widetilde{x}_{ki})+(1+z_{ki})^{2}\right]\\
= & \sum_{k=1}^{K}\widetilde{w}_{k}^{T}D\widetilde{w}_{k}+\sum_{k=1}^{K}\sum_{l=k+1}^{K}\widetilde{w}_{k}^{T}D\widetilde{w}_{l}+\lambda\sum_{k=1}^{K}e^{T}\widetilde{w}_{k}\\
 & +\beta\Biggl(\frac{1}{4}\sum_{k=1}^{K}\sum_{i\in C_{k}}\overset{a_{ki}}{\overbrace{\frac{1}{z_{ki}}}}\widetilde{w}_{k}^{T}\widetilde{x}_{ki}\widetilde{x}_{ki}^{T}\widetilde{w}_{k}-\frac{1}{2}\sum_{k=1}^{K}\sum_{i\in C_{k}}\overset{1+a_{ki}}{\overbrace{(\frac{1+z_{ki}}{z_{ki}})}}(\widetilde{w}_{k}^{T}\widetilde{x}_{ki}) \\ &
 \qquad\qquad\qquad\qquad\qquad\qquad\qquad\qquad+\overset{C}{\overbrace{\sum_{k=1}^{K}\sum_{i\in C_{k}}\frac{1}{4z_{ki}}(1+z_{ki})^{2}}}\Biggr)
\end{aligned}
\end{equation}
where $z_{ki}$ are variables introduced by the majorization scheme into the formulation and $\lambda$ is the Lagrange multiplier. We define the following matrices:

\begin{equation*}
\widetilde{w}\equiv\left[\begin{array}{c}
\widetilde{w}_{1}\\
\widetilde{w}_{2}\\
\vdots\\
\widetilde{w}_{K}
\end{array}\right],
\;U\equiv\left[\begin{array}{c}
e\\
e\\
\vdots\\
e
\end{array}\right]_{L\times1},
\;\widetilde{D}\equiv\left[\begin{array}{cccc}
D & 0 & \cdots & 0\\
0 & D & \cdots & 0\\
\vdots & \vdots & \ddots & \vdots\\
0 & 0 & \cdots & D
\end{array}\right]_{L\times L},\;
\widetilde{D}_{0}\equiv\left[\begin{array}{cccc}
0 & D & \cdots & D\\
D & 0 & \cdots & D\\
\vdots & \vdots & \ddots & \vdots\\
D & D & \cdots & 0
\end{array}\right]_{L\times L},
\end{equation*}
\begin{equation*}
\widetilde{X}\equiv\frac{1}{4}\left[\begin{array}{cccc}
\sum_{i\in C_{1}}a_{1i}\widetilde{x}_{1i}\widetilde{x}_{1i}^{T} & 0 & \cdots & 0\\
0 & \sum_{i\in C_{2}}a_{2i}\widetilde{x}_{2i}\widetilde{x}_{2i}^{T} & \cdots & \vdots\\
\vdots & \vdots & \ddots & 0\\
0 & 0 & \cdots & \sum_{i\in C_{K}}a_{Ki}\widetilde{x}_{Ki}\widetilde{x}_{Ki}^{T}
\end{array}\right]_{L\times L},
\end{equation*}
\begin{equation*}
\widetilde{x}\equiv\frac{1}{2}\left[\begin{array}{c}
\sum_{i\in C_{1}}(1+a_{1i})\widetilde{x}_{1i}\\
\sum_{i\in C_{2}}(1+a_{2i})\widetilde{x}_{2i}\\
\vdots\\
\sum_{i\in C_{K}}(1+a_{Ki})\widetilde{x}_{Ki}
\end{array}\right]_{L\times1}
\end{equation*}
where $L=K(M+1)$, $K$ is the number of classes and $M$ is the number of features. Using these matrices, we rewrite (\ref{extended formulation linear case Lagrangian function}) as a quadratic form
\begin{equation}\label{linear concatinated Lagrangian function}
    \begin{aligned}\mathcal{L}(\widetilde{w},\lambda)=\; & \widetilde{w}^{T}\widetilde{D}\widetilde{w}+\frac{\alpha}{2}\widetilde{w}^{T}\widetilde{D}_{0}\widetilde{w}+\beta\left(\widetilde{w}^{T}\widetilde{X}\widetilde{w}-\widetilde{x}^{T}\widetilde{w}+C\right)+\lambda U^{T}\widetilde{w}\\
=\; & \widetilde{w}^{T}(\widetilde{D}+\frac{\alpha}{2}\widetilde{D}_{0}+\beta\widetilde{X})\widetilde{w}-\beta\widetilde{x}^{T}\widetilde{w}+\beta C+\lambda U^{T}\widetilde{w}.
\end{aligned}
\end{equation}
Now, we differentiate \ref{linear concatinated Lagrangian function} w.r.t. $\widetilde{w}$ to get
\begin{equation*}
    \nabla \mathcal{L}(\widetilde{w})=\frac{\partial \mathcal{L}}{\partial\widetilde{w}}=(2\widetilde{D}+\alpha\widetilde{D}_{0}+2\beta\widetilde{X})\widetilde{w}-\beta\widetilde{x},
\end{equation*}    
\begin{equation*}    
\frac{\partial \mathcal{L}}{\partial\widetilde{w}}=0\:\Longrightarrow2\left(\widetilde{D}+\frac{\alpha}{2}\widetilde{D}_{0}+\beta\widetilde{X}\right)\widetilde{w}=\beta\widetilde{x},
\end{equation*}
\begin{equation*}
\widetilde{w}_{\mathrm{stn}}=\frac{\beta}{2}\left(\widetilde{D}+\frac{\alpha}{2}\widetilde{D}_{0}+\beta\widetilde{X}\right)^{-1}\widetilde{x}.
\end{equation*}

In order for the stationary point $\widetilde{w}_{\mathrm{stn}}$ to be a minimum, the Hessian of $\mathcal{L}$ should be symmetric (which it already is) and positive definite (PD), i.e.,
\begin{equation*}
    H(\widetilde{w})=\frac{\partial \mathcal{L}}{\partial\widetilde{w}\partial\widetilde{w}^{T}}=\widetilde{D}+\frac{\alpha}{2}\widetilde{D}_{0}+\beta\widetilde{X}\succ0.
\end{equation*}
The solution for the $z_{ki}$ follows the approach outlined in Section~\ref{A Quadratic Majorizer for Optimization}. 
\section{Optimization scheme for kernel case with soft constraint on weight vectors and hard constraint on biases}
\label{app_kernel}
Similar to the previous case as explained in Appendix~\ref{app_for_linear}, we rewrite the Lagrangian with soft constraints on the weight vectors ($W$) and hard constraints on the biases ($b$):
\begin{equation*}
   \begin{aligned}\mathcal{L}_{\text{ker}}(W,b,\lambda)= & \sum_{k=1}^{K}\langle w_{k},w_{k}\rangle+\theta\sum_{k=1}^{K}\sum_{l=k+1}^{K}\langle w_{k},w_{l}\rangle\\
 & +\beta\sum_{k=1}^{K}\sum_{i\in C_{k}}\frac{1}{4z_{ki}}\left[1+z_{ki}-(\langle w_{k},\phi(x_{ki})\rangle+b_{k})\right]^{2}+\lambda\sum_{k=1}^{K}b_{k}
\end{aligned}
\end{equation*}
where $\lambda$ is the Lagrange multiplier corresponding to the hard constraint on the biases $\sum_{k=1}^{K}b_{k}=0$. Rewriting the Lagrangian using matrices and using notation introduced in Section~\ref{Kernels} results in
\begin{equation*}
    \begin{aligned}\mathcal{L}_{\text{ker}}(\widetilde{\alpha},\lambda, z)= & \sum_{k=1}^{K}\widetilde{\alpha}_{k}^{T}G_{0}\widetilde{\alpha}_{k}+\theta\sum_{k=1}^{K}\sum_{l=k+1}^{K}\widetilde{\alpha}_{k}^{T}G_{0}\widetilde{\alpha}_{l}\\
 & +\beta\sum_{k=1}^{K}\sum_{i\in C_{k}}\frac{1}{4z_{ki}}\left[1+z_{ki}-(\widetilde{\alpha}_{k}^{T}\widetilde{g}_{ki})\right]^{2}+\lambda\sum_{k=1}^{K}e^{T}\widetilde{\alpha}_{k}
\end{aligned}
\end{equation*}
which is further simplified using matrix notation:
\begin{equation*}
    \begin{aligned}\mathcal{L}_{\text{ker}}(\widetilde{\alpha}, \lambda, z)= &\; \widetilde{\alpha}^{T}\widetilde{G}_{0}\widetilde{\alpha}+\frac{\theta}{2}\widetilde{\alpha}^{T}\widetilde{G}_{1}\widetilde{\alpha}+\beta\left(\widetilde{\alpha}^{T}\widetilde{X}\widetilde{\alpha}-\widetilde{x}^{T}\widetilde{\alpha}+C\right)+\lambda U^{T}\widetilde{\alpha}\\
= &\; \widetilde{\alpha}^{T}(\widetilde{G}_{0}+\frac{\theta}{2}\widetilde{G}_{1}+\beta\widetilde{X})\widetilde{\alpha}+(\lambda U^{T}-\beta\widetilde{x}^{T})\widetilde{\alpha}+\beta C
\end{aligned}
\end{equation*}
where $\widetilde{G}_{0}$ ,$\widetilde{G}_{1}$ and $U$ are necessary matrices for rewriting Lagrangian after concatenating the $\widetilde{\alpha}_{k}$. From the KKT necessary conditions, we get
\begin{equation*}
    \frac{\partial \mathcal{L}_{\text{ker}}}{\partial\widetilde{\alpha}}=0\:\Longrightarrow2\left(\widetilde{G}_{0}+\frac{\theta}{2}\widetilde{G}_{1}+\beta\widetilde{X}\right)\widetilde{\alpha}+\lambda U=\beta\widetilde{x}
\end{equation*}
\begin{equation*}
\frac{\partial \mathcal{L}_{\text{ker}}}{\partial\lambda}=0\:\Longrightarrow U^{T}\widetilde{\alpha}=0
\end{equation*}
Rewriting these equations in matrix format, we get
\begin{equation*}
\left[\begin{array}{cc}
2\widetilde{G}_{0}+\theta\widetilde{G}_{1}+2\beta\widetilde{X} & U\\
U^{T} & 0
\end{array}\right]\left[\begin{array}{c}
\widetilde{\alpha}\\
\lambda
\end{array}\right]=\left[\begin{array}{c}
\beta\widetilde{x}\\
0.
\end{array}\right]    
\end{equation*}
The solution for the $z_{ki}$ follows the approach outlined in Section~\ref{A Quadratic Majorizer for Optimization}.
After obtaining $\left\{\widetilde{\alpha}_{1},\widetilde{\alpha}_{2},\cdots,\widetilde{\alpha}_{K}\right\}$, for each class $k$, we calculate the projection in RKHS for a test pattern as follows:
\begin{equation*}
    \begin{aligned}\ensuremath{\langle w_{k},\phi(x_{t})-x_{0}\rangle}= & \langle\sum_{i\in SV}\alpha_{ki}\phi(x_{i}),\phi(x_{t})\rangle+b_{k}\\
= & \sum_{i\in SV}\alpha_{ki}\langle\phi(x_{i}),\phi(x_{t})\rangle+b_{k}\\
= & \sum_{i\in SV}\alpha_{ki}\langle\phi(x_{i}),\phi(x_{t})\rangle+b_{k}\\
= & \sum_{i\in SV}\alpha_{ki}K(x_{i},x_{t})+b_{k}\\
= &\; \widetilde{\alpha}_{k}^{T}g_{t}^{\mathrm{Test}}
\end{aligned}
\end{equation*}
where $g_{t}^{\mathrm{Test}}$ is the $t$th column of the Gram matrix between training support vectors and test patterns.
\newpage
\vskip 0.2in
\bibliographystyle{abbrv}
\bibliography{SVM.bib}

\end{document}